\documentclass{article}

\usepackage{iclr2027_conference,times}
\usepackage{url}
\usepackage{booktabs}
\usepackage{graphicx}
\usepackage{amsmath,amssymb,mathtools}
\usepackage{microtype}
\usepackage{xcolor}
\usepackage{multirow}
\usepackage{tabularx}
\usepackage{array}
\usepackage{capt-of}
\usepackage{placeins}
\usepackage{algorithm}
\usepackage{algpseudocode}
\usepackage{tikz}
\usetikzlibrary{arrows.meta,positioning,fit,calc,decorations.pathreplacing}
\usepackage{hyperref}
\ifdefined\SCFFMainOnly
\usepackage{xr}
\fi

\definecolor{scffblue}{HTML}{2F6690}
\definecolor{scffgreen}{HTML}{3A7D6B}
\definecolor{scffamber}{HTML}{A66A18}
\definecolor{scffgray}{HTML}{66717E}
\definecolor{scffred}{HTML}{A3473F}
\newcolumntype{Y}{>{\raggedright\arraybackslash}X}
\newcommand{\scff}{\textsc{SCFF}}
\newcommand{\algphase}[2]{%
  \Statex\colorbox{#1!13}{\parbox{\dimexpr\linewidth-2\fboxsep\relax}{%
    \textcolor{#1}{\textbf{#2}}}}}

\hypersetup{
  pdftitle={Support-Compiled Feature Folding: More Evidence at Lower Memory Across Tabular Foundation Models},
  pdfauthor={Tian Zhou, Beverly Jin, Xue Wang, Linxiao Yang, Wenwei Wang, Bingqing Peng, Mengni Ye, Jinjie Gu, and Liang Sun},
  colorlinks=true,
  linkcolor=scffblue,
  citecolor=scffblue,
  urlcolor=scffblue
}

\title{Support-Compiled Feature Folding:\\
More Evidence at Lower Memory\\
Across Tabular Foundation Models}
\author{\normalfont
Tian Zhou\textsuperscript{1,\textdagger},
Beverly Jin\textsuperscript{2,\textdagger},
Xue Wang\textsuperscript{2},
Linxiao Yang\textsuperscript{1},
Wenwei Wang\textsuperscript{1}\\
Bingqing Peng\textsuperscript{1},
Mengni Ye\textsuperscript{2},
Jinjie Gu\textsuperscript{1},
Liang Sun\textsuperscript{1}\\[-0.15em]
\small\textsuperscript{1}Ant Group
\qquad
\textsuperscript{2}Independent Researcher\\[-0.15em]
\small\textsuperscript{\textdagger}Equal contribution
}

\iclrfinalcopy

\begin{document}
\raggedbottom
\maketitle
\lhead{Submitted to ICLR 2027}

\begin{abstract}
Wide tables offer tabular foundation models more evidence, but accessing it
can exhaust their memory: full-width pairwise mixing grows quadratically with
the number of columns, while feature selection makes inputs affordable by
discarding evidence. We ask whether using more features requires interacting
over all of them at once. We introduce
Support-Compiled Feature Folding (\scff{}), a training-free inference
framework that encodes wide tables through bounded calls to a frozen backbone.
SCFF organizes support-ranked features into a strong Core and a candidate Tail,
folds them into narrow feature groups, and support-checks the Tail's added
evidence before a single contextual prediction. This converts
quadratic feature-interaction work into linear-in-width work with a bounded
local working set, without ensembling predictions or training new parameters.

On the exhaustive 18-dataset wide-table slice of fixed AMLB-29, TabZilla, and
TabArena snapshots, SCFF improves dataset-macro accuracy and NLL on all six
evaluated backbones. All four matched-width comparisons retain favorable 95\%
dataset-bootstrap intervals on locked folds, with relative error reductions
up to $26.1\%$. Median paired GPU-memory savings are $2.09$--$2.36\times$,
and the ratio of separately observed maximum peaks reaches $34.3\times$.
Under a measured peak-memory ceiling, SCFF uses the saved budget to preserve
more support-selected evidence, improving accuracy by $4.06$ and $3.72$
points over the widest feasible single leaf on predeclared wide-Core strata
of TabICLv2 and TabPFN-3.
\end{abstract}

\section{Introduction}

Tabular foundation models promise to turn labeled examples into a predictor
without task-specific training
\citep{mueller2022pfn,hollmann2025tabpfn,qu2025tabicl}. On a wide table,
however, even presenting the available evidence can be expensive. Thousands
of columns offer many possible clues, but their usefulness is uneven: some
carry a clear signal, others matter only together, and many are redundant.
Full-width feature attention pays for their pairwise interactions before the
model can reason across examples. Feature selection makes this affordable by
discarding columns, including any useful evidence beyond its cutoff.
The resulting dilemma is concrete: \emph{how much evidence must a frozen
model give up simply to fit the table in memory?}

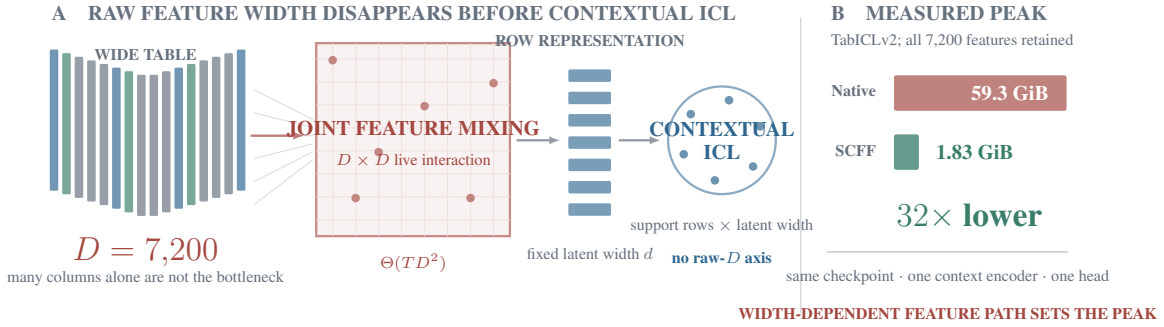
\begin{figure}[t]
\centering
\resizebox{\linewidth}{!}{
\begin{tikzpicture}[x=1cm,y=1cm,
 note/.style={font=\footnotesize,text=scffgray,align=center},
 flow/.style={-{Latex[length=1.7mm]},line width=1.0pt}]
\path[use as bounding box] (0,0) rectangle (16,5.45);

\node[note,font=\small\bfseries,anchor=west] at (0.15,5.12)
 {A\quad RAW FEATURE WIDTH DISAPPEARS BEFORE CONTEXTUAL ICL};

\foreach \i in {0,...,15}{
 \pgfmathtruncatemacro{\kind}{mod(\i,5)}
 \ifnum\kind=0\def\cc{scffblue}\else\ifnum\kind=1\def\cc{scffgreen}\else\def\cc{scffgray}\fi\fi
 \fill[\cc!68,rounded corners=0.6pt]
   ({0.24+0.19*\i},{2.00+0.055*abs(7.5-\i)}) rectangle ++(0.13,2.15);
}
\node[note,font=\scriptsize\bfseries] at (1.70,4.48) {WIDE TABLE};
\node[font=\Large\bfseries,text=scffred] at (1.70,1.48) {$D=7{,}200$};
\node[note,font=\scriptsize] at (1.70,1.08) {many columns alone are not the bottleneck};

\foreach \y in {2.20,2.55,...,4.30}{
 \draw[scffgray!25,line width=0.35pt] (3.35,\y)--(4.25,{3.25+0.35*(\y-3.25)});
}
\draw[flow,scffred!70] (3.30,3.25)--(4.12,3.25);
\fill[scffred!7] (4.30,1.72) rectangle (7.25,4.66);
\draw[scffred!62,line width=1.0pt] (4.30,1.72) rectangle (7.25,4.66);
\foreach \x in {4.55,4.90,...,7.05}{
 \draw[scffred!16,line width=0.35pt] (\x,1.72)--(\x,4.66);
 \draw[scffred!16,line width=0.35pt] (4.30,{\x-2.58})--(7.25,{\x-2.58});
}
\foreach \x/\y in {4.55/4.40,5.25/3.00,5.95/3.70,6.65/2.30,7.00/4.05,4.90/2.30}
 \fill[scffred!66] (\x,\y) circle (0.065);
\node[note,font=\footnotesize\bfseries,text=scffred] at (5.78,3.35)
 {JOINT FEATURE MIXING};
\node[note,font=\scriptsize,text=scffred] at (5.78,2.88) {$D\times D$ live interaction};
\node[note,font=\scriptsize\bfseries,text=scffred] at (5.78,1.32) {$\Theta(TD^2)$};

\draw[flow,scffgray!70] (7.35,3.18)--(8.02,3.18);
\foreach \i in {0,...,6}
 \fill[scffblue!64,rounded corners=0.7pt] (8.14,{2.03+0.34*\i}) rectangle ++(0.65,0.19);
\node[note,font=\scriptsize\bfseries] at (8.47,4.70) {ROW REPRESENTATION};
\node[note,font=\scriptsize] at (8.47,1.45) {fixed latent width $d$};

\draw[flow,scffgray!70] (8.91,3.18)--(9.55,3.18);
\draw[scffblue!55,line width=1.0pt] (10.48,3.18) circle (0.82);
\foreach \a in {20,80,...,320}{
 \fill[scffblue!70] ({10.48+0.62*cos(\a)},{3.18+0.62*sin(\a)}) circle (0.065);
}
\node[note,font=\footnotesize\bfseries,text=scffblue] at (10.48,3.18)
 {CONTEXTUAL\\ICL};
\node[note,font=\scriptsize] at (10.48,1.87) {support rows $\times$ latent width};
\node[note,font=\scriptsize\bfseries,text=scffblue] at (10.48,1.40) {no raw-$D$ axis};

\draw[scffgray!28,line width=0.6pt] (11.68,0.55)--(11.68,5.05);
\node[note,font=\small\bfseries,anchor=west] at (12.02,5.12)
 {B\quad MEASURED PEAK};
\node[note,font=\scriptsize,anchor=west] at (12.02,4.70)
 {TabICLv2; all 7,200 features retained};

\node[note,font=\scriptsize\bfseries,anchor=east] at (12.95,3.93) {Native};
\fill[scffred!68,rounded corners=1pt] (13.10,3.63) rectangle (15.73,4.17);
\node[font=\footnotesize\bfseries,text=white,anchor=east] at (15.58,3.90) {59.3 GiB};

\node[note,font=\scriptsize\bfseries,anchor=east] at (12.95,3.03) {SCFF};
\fill[scffgreen!78,rounded corners=1pt] (13.10,2.73) rectangle (13.48,3.27);
\node[font=\footnotesize\bfseries,text=scffgreen,anchor=west] at (13.62,3.00) {1.83 GiB};

\node[font=\Large\bfseries,text=scffgreen] at (14.25,2.02) {$32\times$ lower};
\draw[scffgray!45,line width=0.55pt] (12.08,1.48)--(15.75,1.48);
\node[note,font=\scriptsize] at (13.92,1.10)
 {same checkpoint $\cdot$ one context encoder $\cdot$ one head};
\node[note,font=\scriptsize\bfseries,text=scffred] at (13.92,0.54)
 {WIDTH-DEPENDENT FEATURE PATH SETS THE PEAK};
\end{tikzpicture}}
\caption{\textbf{Feature width is a distinct bottleneck in tabular ICL.}
\textbf{A:} Width-dependent interactions precede the fixed-dimensional row
representation consumed by contextual ICL. \textbf{B:}
An illustrative TabICLv2 folding pilot retains all 7,200 features of
\emph{robert} in leaves of at most 128 columns, reducing peak allocated GPU
memory from $59.3$ to $1.83$ GiB with the same downstream stack and head.
Table~\ref{tab:full-coverage-main} evaluates the final SCFF method separately.}
\label{fig:width-problem}
\end{figure}

Consider \emph{robert}, with 7,200 features. In an illustrative all-feature
folding pilot, TabICLv2 uses $59.3$ GiB of peak allocated GPU memory. Processing
the same columns in groups of at most 128 lowers the peak to $1.83$ GiB while
keeping the checkpoint, context encoder, and prediction head unchanged
(Figure~\ref{fig:width-problem}). This gap motivates a different question:
\emph{does using a wide table require interacting over its full width at once?}

The opportunity lies at an existing boundary inside the model. In
feature-interaction ICL architectures, a feature encoder compresses each row
into a fixed-dimensional representation; the context encoder then reasons
across rows after the raw feature axis has disappeared. The costly pairwise
mixing, with $\Theta(TD^2)$ interactions for $T$ rows and $D$ features, occurs
before this boundary. Our key insight is that \emph{the feature encoder can
see a narrow group while the context encoder still sees every row}.
Multiple groups can contribute to the same row representation, allowing the
downstream model to make one prediction from their combined evidence.

We turn this observation into \emph{Support-Compiled Feature Folding}
(\scff{}). Support labels organize the columns into a Core of individually
informative features and a candidate Tail. Each is divided into bounded
groups, or \emph{leaves}, processed by the same frozen feature encoder.
SCFF merges their row representations before the context encoder. With fixed
leaf capacity, feature-interaction work grows linearly with total width,
while all support examples remain available to the final prediction.

Making evidence cheaper to encode is only half the problem: adding more
evidence can also add redundancy and noise. SCFF therefore gives the Tail a
specific role---to contribute information that the Core has not already
captured. It removes the Core-aligned component and uses reciprocal tests on
support subsets to check whether the residual contributes reproducible
evidence. An accepted residual augments the Core; rejection preserves its
representation exactly. Every column is encoded, but Tail inclusion is
conditional. Binary tasks use this support-checked update; multiclass tasks
use Bounded Core. A token-preserving LimiX adapter extends the design to a
different model interface.

The practical payoff is more than fitting a wide input. Memory spent on
feature interactions competes with the budget for retaining features and
providing labeled context. Just as retrieval improves the value of context
slots \citep{thomas2024retrieval}, folding can improve how the model spends
its feature-side budget. We test this on a fixed OpenML Wide-Table benchmark:
all eligible dataset versions in the AMLB-29, TabZilla, and TabArena snapshots
with at least 128 raw features, under declared class and development-exclusion
rules. Its 18 datasets let us ask whether the design transfers across frozen
models, whether folding helps beyond selection, and whether the saved memory
can buy useful evidence. Our contributions are:
\begin{itemize}
 \item \textbf{A training-free width compiler.} We introduce \scff{}, which
 combines support-conditioned routing, bounded feature folding, and
 support-checked residual evidence before one context-encoder and label-head
 pass. It trains no parameters and does not ensemble predictions.
 \item \textbf{Broad gains on wide tables.} Across six frozen backbones on the
 exhaustive 18-dataset wide-table slice, SCFF improves every dataset-macro
 accuracy and NLL comparison. Relative error falls by up to $26.1\%$, relative
 NLL by up to $19.0\%$, and the maximum-peak memory ratio reaches $34.3\times$.
 On locked folds, all four matched-width backbones retain favorable 95\%
 dataset-bootstrap intervals for both predictive metrics.
 \item \textbf{A mechanism for reinvesting memory in evidence.} Holding the
 selected Core fixed, bounded leaves reduce peak memory across three
 latent-state backbones. Under the measured folded-memory ceiling, SCFF keeps
 more selected features than the widest feasible single leaf and gains $4.06$
 and $3.72$ accuracy points on the predeclared wide-Core strata of TabICLv2
 and TabPFN-3. Support-checked Tail recovery then tests evidence outside the
 Core before the same single prediction.
\end{itemize}

\section{Related Work}

\paragraph{Learning a predictor, adapting its input.}
Prior-data fitted networks amortize Bayesian-style inference over synthetic
tasks, predicting from a labeled support table without dataset-specific
gradient updates \citep{mueller2022pfn}. TabPFN and its successors improve the
prior, tokenizer, and inference stack
\citep{hollmann2025tabpfn,grinsztajn2025tabpfn25,grinsztajn2026tabpfn3}, while
TabICL scales the paradigm toward larger tables
\citep{qu2025tabicl,qu2026tabiclv2}. Released systems such as LimiX, EXAONE
Tabular, and TabFM broaden the family of pretrained tabular predictors
\citep{limix2025,eo2026exaone,tabfmcode2026}. These advances make a frozen
predictor a useful starting point. SCFF asks how much wider an input it can
use through inference-time organization alone, exploiting the interface
between feature encoding and contextual prediction.

\paragraph{Making more evidence affordable.}
Retrieval allocates finite context to relevant support examples
\citep{thomas2024retrieval}. We address the complementary feature-width
budget. Classical filters rank or delete columns using support
statistics \citep{guyon2003variable}, with multiple-testing control providing
a principled high-evidence subset \citep{benjamini1995controlling}; feature
engineering instead changes the input schema \citep{tschalzev2026tabprep}.
Test-time divide-and-conquer extends TabPFNv2 without retraining
\citep{ye2025closer}, while TabPFN-Wide uses continued pre-training to handle
extreme feature counts \citep{kolberg2025wide}. SCFF explores a complementary
choice: retain the frozen predictor and reorganize its feature computation.
It uses ANOVA/BH to identify a Core, encodes the remaining columns as a
candidate Tail, and combines intermediate representations before one
contextual prediction. This locates the width adaptation inside feature
encoding, where the pairwise cost arises.

\paragraph{Bounded computation, shared prediction.}
Deep Sets and Set Transformer learn permutation-invariant aggregation end to
end \citep{zaheer2017deep,lee2019set}. SCFF instead reuses one feature
encoder on disjoint leaves and combines its row representations without a learned set
encoder. This resembles the bounded-interface principle of hierarchical
classification \citep{silla2011hierarchical,qu2025tabicl}, but feature groups
produce row representations rather than predictions. They are merged before the
shared context encoder, which retains all support and query rows, and the
label head executes once. The relevant design principle is to expand the
evidence handled by a model through repeated use of a bounded interface.
For features, the shared object is the representation on which the model
will reason about the task.

\section{Support-Compiled Feature Folding}
\label{sec:method}

The design begins with a separation: a model can collect evidence in narrow
groups and still reason over all examples together. SCFF uses the boundary
between feature encoding and contextual prediction to implement this
separation. For an episode $(X_s,y_s,X_q)$ with $D$ features and $T$
support-plus-query rows, write
\begin{equation}
  f_\theta(X_s,y_s,X_q)
  =H_\theta\!\left(C_\theta\!\left(
  F_\theta([X_s;X_q],y_s);y_s\right)\right),
  \label{eq:model-interface}
\end{equation}
where $F_\theta$ is the width-dependent feature encoder,
$C_\theta$ is the context encoder, and $H_\theta$ is the label head.
We keep all pretrained parameters fixed. SCFF changes how $F_\theta$ is
executed and leaves the remaining model layers unchanged.

\begin{figure}[t]
\centering
\resizebox{\linewidth}{!}{
\begin{tikzpicture}[x=1cm,y=1cm,
 note/.style={font=\scriptsize,text=scffgray,align=center},
 smallnote/.style={font=\tiny,text=scffgray,align=center},
 head/.style={font=\scriptsize\bfseries,text=scffgray,align=center},
 flow/.style={-{Latex[length=1.5mm]},line width=0.9pt,draw=scffgray!68},
 panel/.style={rounded corners=6pt,line width=0.9pt}]
\path[use as bounding box] (0,-0.05) rectangle (16,8.45);

\node[head,anchor=west,text=scffblue] at (0.18,8.18)
 {A\quad COMPILE ON LABELED SUPPORT ROWS};
\node[smallnote,anchor=east] at (15.82,8.18)
 {no query label, no learned parameter};
\draw[scffblue!28,line width=0.7pt] (0.18,7.96)--(15.82,7.96);

\draw[panel,draw=scffblue!52,fill=scffblue!3] (0.18,4.70) rectangle (4.35,7.70);
\node[head,text=scffblue] at (2.265,7.42) {1. ROUTE BY EVIDENCE};
\foreach \r in {0,...,5}{\foreach \c in {0,...,8}{
 \pgfmathtruncatemacro{\kind}{mod(2*\c+\r,5)}
 \ifnum\kind=0\def\cc{scffblue}\else\ifnum\kind=1\def\cc{scffgreen}\else\def\cc{scffgray}\fi\fi
 \fill[\cc!52] (0.46+0.15*\c,6.08+0.14*\r) rectangle ++(0.11,0.10);
}}
\node[smallnote] at (1.14,5.86) {support rows};
\foreach \r in {0,...,5}{
 \pgfmathtruncatemacro{\kind}{mod(\r,2)}
 \ifnum\kind=0\def\cc{scffblue}\else\def\cc{scffamber}\fi
 \fill[\cc!85] (1.97,6.13+0.14*\r) circle (0.042);
}
\node[smallnote] at (1.97,5.86) {$y_s$};
\draw[flow] (2.14,6.48)--(2.43,6.48);
\draw[scffgray!32,line width=0.5pt] (2.58,5.95)--(2.58,6.94)--(4.02,6.94);
\foreach \x/\h/\cc in {0/0.22/scffgray,1/0.78/scffblue,2/0.31/scffgray,3/0.91/scffblue,4/0.18/scffgray,5/0.61/scffblue,6/0.26/scffgray}{
 \fill[\cc!72] (2.72+0.17*\x,5.97) rectangle ++(0.105,\h);
}
\node[smallnote] at (3.32,5.72) {ANOVA + BH};
\draw[rounded corners=3pt,draw=scffblue!60,fill=scffblue!10]
 (0.48,5.02) rectangle (2.05,5.48);
\fill[scffblue!78] (0.65,5.17) circle (0.07);
\node[smallnote,text=scffblue,anchor=west,font=\tiny\bfseries] at (0.82,5.25) {CORE};
\node[smallnote,anchor=west] at (0.82,5.09) {high evidence};
\draw[rounded corners=3pt,draw=scffgray!50,fill=white]
 (2.23,5.02) rectangle (4.05,5.48);
\fill[scffgray!62] (2.40,5.17) circle (0.07);
\node[smallnote,anchor=west,font=\tiny\bfseries] at (2.57,5.25) {TAIL};
\node[smallnote,anchor=west] at (2.57,5.09) {remaining evidence};

\draw[flow] (4.38,6.20)--(4.68,6.20);
\draw[panel,draw=scffgreen!55,fill=scffgreen!3] (4.73,4.70) rectangle (10.15,7.70);
\node[head,text=scffgreen] at (7.44,7.42) {2. FOLD EVERY COLUMN};
\foreach \i in {0,...,15}{
 \pgfmathtruncatemacro{\kind}{mod(3*\i+1,7)}
 \ifnum\kind<2\def\cc{scffblue}\else\def\cc{scffgray}\fi
 \draw[rounded corners=1pt,draw=\cc!58,fill=\cc!12]
 (5.05+0.29*\i,6.78) rectangle ++(0.23,0.32);
}
\node[smallnote,anchor=west] at (5.05,6.57) {complete feature axis};
\node[smallnote,text=scffgreen,anchor=east,font=\tiny\bfseries] at (9.89,6.57)
 {EACH LEAF $\le M$};
\draw[flow] (7.44,6.47)--(7.44,6.12);
\draw[scffblue!65,line width=0.85pt] (7.44,6.12)--(6.37,5.90);
\draw[scffgray!62,line width=0.85pt] (7.44,6.12)--(8.52,5.90);
\node[smallnote,text=scffblue,font=\tiny\bfseries] at (5.95,6.09) {CORE LEAVES};
\node[smallnote,font=\tiny\bfseries] at (8.91,6.09) {TAIL LEAVES};
\foreach \x in {0,1}{
 \draw[rounded corners=3pt,draw=scffblue!68,fill=scffblue!10]
 (5.22+0.92*\x,5.18) rectangle ++(0.76,0.58);
 \foreach \k in {0,...,3}{\fill[scffblue!70]
   (5.33+0.92*\x+0.14*\k,5.38) rectangle ++(0.09,0.18);}
}
\foreach \x in {0,1,2}{
 \draw[rounded corners=3pt,draw=scffgray!55,fill=white]
 (7.35+0.80*\x,5.18) rectangle ++(0.65,0.58);
 \foreach \k in {0,...,3}{\fill[scffgray!52]
   (7.45+0.80*\x+0.12*\k,5.38) rectangle ++(0.075,0.18);}
}
\node[smallnote,text=scffgreen] at (7.44,4.93) {all columns are encoded};

\draw[flow] (10.18,6.20)--(10.48,6.20);
\draw[panel,draw=scffamber!62,fill=scffamber!3] (10.53,4.70) rectangle (15.82,7.70);
\node[head,text=scffamber] at (13.175,7.42) {3. OPTIONAL TAIL GATE};
\node[circle,draw=scffblue!70,fill=scffblue!10,minimum size=6.5mm,
 text=scffblue,font=\scriptsize\bfseries] at (11.40,6.47) {$c$};
\node[circle,draw=scffgray!65,fill=white,minimum size=6.5mm,
 text=scffgray,font=\scriptsize\bfseries] at (11.40,5.68) {$t$};
\draw[flow] (11.78,6.08)--(12.43,6.08);
\node[circle,draw=scffamber!75,fill=scffamber!10,minimum size=8mm,
 text=scffamber,font=\scriptsize\bfseries] at (12.82,6.08) {$r\!\perp\!c$};
\node[smallnote] at (12.82,5.38) {remove Core projection};
\draw[flow] (13.26,6.08)--(13.71,6.45);
\draw[flow] (13.26,6.08)--(13.71,5.72);
\draw[rounded corners=3pt,draw=scffgray!55,fill=white]
 (13.75,6.20) rectangle (14.70,6.70);
\node[smallnote,font=\tiny\bfseries] at (14.225,6.45) {RAW};
\draw[rounded corners=3pt,draw=scffgreen!55,fill=scffgreen!7]
 (13.75,5.46) rectangle (14.70,5.96);
\node[smallnote,text=scffgreen,font=\tiny\bfseries] at (14.225,5.71) {PROTOTYPE};
\draw[flow] (14.73,6.45)--(15.18,6.18);
\draw[flow] (14.73,5.71)--(15.18,5.99);
\draw[rounded corners=3pt,draw=scffamber!65,fill=white]
 (15.18,5.63) rectangle (15.66,6.52);
\node[smallnote,text=scffamber,font=\tiny\bfseries,rotate=90] at (15.42,6.08)
 {A $\leftrightarrow$ B};
\node[smallnote] at (13.18,4.93) {Fisher + reciprocal acceptance test};

\node[head,anchor=west,text=scffgreen] at (0.18,4.34)
 {B\quad RUN THE MODEL};
\node[smallnote,anchor=east] at (15.82,4.34)
 {bounded Core $+$ accepted residual $\rightarrow$ one prediction};
\draw[scffgreen!28,line width=0.7pt] (0.18,4.12)--(15.82,4.12);
\draw[panel,draw=scffgreen!45,fill=scffgreen!1] (0.18,0.18) rectangle (15.82,3.86);

\node[head,text=scffgray] at (1.25,3.48) {BOUNDED LEAVES};
\foreach \y/\lab in {3.03/$C_1$,2.57/$C_2$}{
 \draw[rounded corners=2pt,draw=scffblue!65,fill=scffblue!10]
 (0.45,\y-0.17) rectangle (1.12,\y+0.17);
 \node[smallnote,text=scffblue,font=\tiny\bfseries] at (0.785,\y) {\lab};
 \draw[flow,scffblue] (1.14,\y)--(1.65,\y);
}
\foreach \y/\lab in {1.91/$T_1$,1.51/$T_2$,1.11/$T_3$,0.71/$\cdots$}{
 \draw[rounded corners=2pt,draw=scffgray!52,fill=white]
 (0.45,\y-0.15) rectangle (1.12,\y+0.15);
 \node[smallnote,font=\tiny\bfseries] at (0.785,\y) {\lab};
 \draw[flow] (1.14,\y)--(1.65,\y);
}
\draw[rounded corners=4pt,draw=scffblue!65,fill=white]
 (1.69,0.48) rectangle (2.75,3.25);
\node[note,text=scffblue,font=\scriptsize\bfseries] at (2.22,2.05) {shared};
\node[note,text=scffblue,font=\scriptsize\bfseries] at (2.22,1.70) {$F_\theta$};
\node[smallnote] at (2.22,1.24) {same feature};
\node[smallnote] at (2.22,1.04) {encoder};
\foreach \y in {2.92,2.55}{\draw[flow,scffblue] (2.77,\y)--(3.28,\y);}
\foreach \y in {1.86,1.48,1.10,0.72}{\draw[flow] (2.77,\y)--(3.28,\y);}

\draw[rounded corners=5pt,draw=scffblue!68,fill=scffblue!10]
 (3.32,2.28) rectangle (7.16,3.18);
\node[head,text=scffblue] at (5.24,2.91) {BOUNDED CORE};
\node[note,text=scffblue,font=\scriptsize\bfseries] at (5.24,2.55)
 {$c=\operatorname{mean}(m_{C_g})$};

\draw[rounded corners=5pt,draw=scffgray!52,fill=white]
 (3.32,0.47) rectangle (5.38,1.98);
\node[smallnote,font=\tiny\bfseries] at (4.35,1.72) {TAIL SUMMARY};
\foreach \x/\y in {3.66/1.24,4.02/1.06,4.33/1.35,4.67/1.12,5.02/1.42}{
 \fill[scffgray!58] (\x,\y) circle (0.075);
}
\node[smallnote] at (4.35,0.70) {$t=\operatorname{mean}(m_{T_g})$};
\draw[flow] (5.41,1.22)--(5.78,1.22);
\draw[rounded corners=5pt,draw=scffamber!62,fill=scffamber!6]
 (5.82,0.47) rectangle (8.05,1.98);
\node[smallnote,text=scffamber,font=\tiny\bfseries] at (6.935,1.72) {TAIL RESIDUAL};
\node[note,text=scffamber,font=\scriptsize\bfseries] at (6.935,1.25) {$r=t-\rho c$};
\node[smallnote] at (6.935,0.76) {raw or prototype map};
\draw[flow] (8.08,1.22)--(8.47,1.22);
\draw[rounded corners=5pt,draw=scffgreen!60,fill=scffgreen!6]
 (8.51,0.47) rectangle (10.75,1.98);
\node[smallnote,text=scffgreen,font=\tiny\bfseries] at (9.63,1.72) {SUPPORT GATE};
\node[note,text=scffgreen,font=\scriptsize\bfseries] at (9.63,1.27)
 {$\alpha v$ or $0$};
\node[smallnote] at (9.63,0.76) {add only accepted signal};

\draw[flow,scffblue,line width=1.2pt] (7.19,2.73)--(11.10,2.16);
\draw[flow,scffgreen] (10.78,1.22)--(11.10,1.78);
\node[circle,draw=scffgreen!75,fill=scffgreen!10,minimum size=10mm,
 text=scffgreen,font=\scriptsize\bfseries] at (11.45,1.96) {$+$};
\node[smallnote,text=scffgreen,font=\tiny\bfseries] at (11.45,0.62)
 {$h=c+\alpha v$};

\draw[flow,scffgreen] (11.92,1.96)--(12.32,1.96);
\draw[rounded corners=4pt,draw=scffgreen!62,fill=scffgreen!5]
 (12.36,1.25) rectangle (14.32,2.67);
\node[head,text=scffgreen] at (13.34,2.35) {ORIGINAL MODEL};
\node[note,text=scffblue,font=\scriptsize\bfseries] at (13.34,1.91)
 {$C_\theta\;\rightarrow\;H_\theta$};
\node[smallnote] at (13.34,1.53) {one stack, one head};
\draw[flow,scffgreen] (14.35,1.96)--(14.83,1.96);
\foreach \i/\hh in {0/0.24,1/0.53,2/0.82}{
 \fill[scffgreen!78] (14.92+0.15*\i,1.45) rectangle ++(0.085,\hh);
}
\node[smallnote] at (15.15,1.11) {one prediction};

\end{tikzpicture}}
\caption{\textbf{SCFF folds the Core and optionally recovers Tail evidence.}
Support labels route every column to Core or Tail, and the same feature
encoder processes each bounded-width leaf. Core leaves capture the strongest
support evidence without full-width competition. Tail leaves propose
nonredundant evidence, which enters the prediction only after a reciprocal
support-subset acceptance test. The combined row representation enters the original
context encoder and label head once.}
\label{fig:scff}
\end{figure}
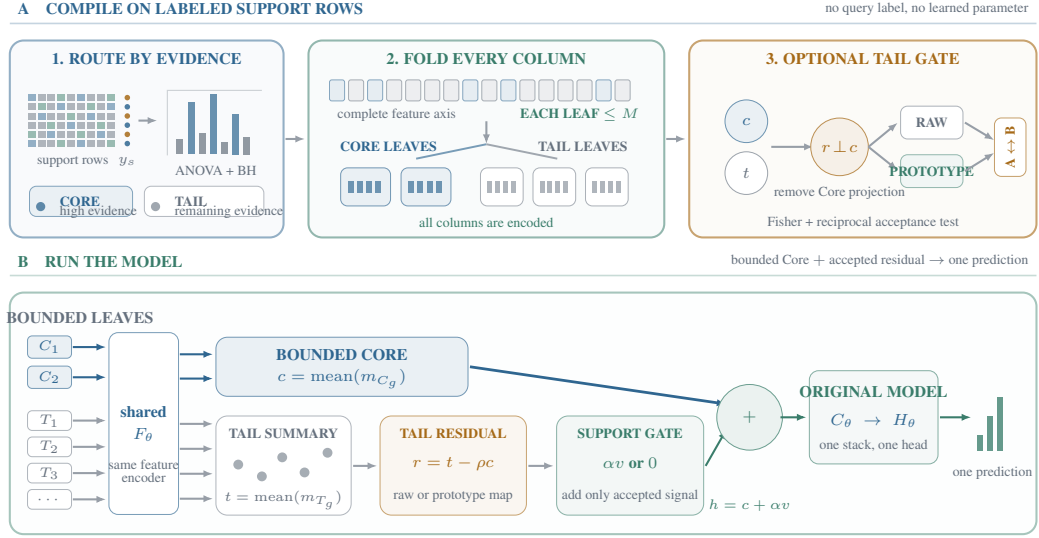

\subsection{Organize evidence before compressing it}

Treating all columns equally would mix strong individual signals with weak
or redundant evidence before either is understood. Algorithm~\ref{alg:scff}
first ranks columns by one-way ANOVA: class separation relative to within-class
variation on a median-filled support copy. The Benjamini--Hochberg discovery
count $K$ determines Core size; the strongest $\max(1,K)$ columns form
$\mathcal C$, and the rest form the candidate Tail $\mathcal T$. Weak
individual scores need not imply useless joint information, so the Tail is
encoded for a later check. Routing reads support labels only; the model
retains its native preprocessing and missing-value handling.

Within each stratum, round-robin assignment creates the minimum number of
disjoint leaves of at most $M=128$ columns. Every preprocessed feature occurs
in exactly one leaf. The same feature encoder maps each leaf to a row-aligned
representation
\begin{equation}
  m_g=F_\theta([X_s;X_q]_{P_g},y_s)
  \in\mathbb R^{T\times d_{\rm ICL}}.
  \label{eq:leaf-message}
\end{equation}
Each leaf therefore preserves every support and query row in the same latent
coordinates. This is the point where many narrow feature computations become
compatible inputs to one contextual prediction.

\begin{algorithm}[t]
\caption{Support-Compiled Feature Folding (row-representation adapter)}
\label{alg:scff}
\small
\textbf{Input:} episode $(X_s,y_s,X_q)$; model
$(F_\theta,C_\theta,H_\theta)$; leaf capacity $M$\hfill
\textbf{Output:} one query distribution
\par\smallskip
\begin{minipage}[t]{0.485\linewidth}
\begin{algorithmic}
\algphase{scffamber}{ROUTE --- organize every column}
\State Score columns by support-only ANOVA
\State $K\gets$ BH discovery count at FDR $0.05$
\State $\mathcal C\gets$ top $\max(1,K)$; $\mathcal T\gets[D]\setminus\mathcal C$
\State Partition each stratum round-robin into
\Statex\hspace{\algorithmicindent}disjoint leaves $P_g$, $|P_g|\le M$

\algphase{scffblue}{FOLD + ENCODE --- reuse the feature encoder}
\For{each Core or Tail leaf $P_g$}
  \State $m_g\gets F_\theta([X_s;X_q]_{P_g},y_s)$
\EndFor

\algphase{scffblue}{CORE --- preserve strong evidence}
\State $c\gets |\mathcal G_C|^{-1}\sum_{g\in\mathcal G_C}m_g$
\end{algorithmic}
\end{minipage}\hfill
\begin{minipage}[t]{0.485\linewidth}
\begin{algorithmic}
\algphase{scffgreen}{OPTIONAL TAIL --- test added information}
\State $\alpha\gets0$; $v\gets0$
\If{$\mathcal T\neq\varnothing$ and the task is binary}
  \State Average Tail messages and center on support
  \State Remove the Core-aligned projection
  \State Form normalized raw and prototype candidates
  \State Split every support class into halves $A,B$
  \For{each candidate $v'$}
    \State Fit Fisher steps on $A$ and $B$
    \State Test each step on the opposite half
    \State Require NLL and separation gains on $A,B$
    \State Require no accuracy loss on $A,B$
  \EndFor
  \State Select the accepted update with the largest
  \Statex\hspace{\algorithmicindent}worst-direction NLL gain
  \State For one Tail leaf, allow Fisher-only validation
\EndIf

\algphase{scffgray}{PREDICT --- use the original context encoder}
\State \Return $H_\theta(C_\theta(c+\alpha v;y_s))$
\end{algorithmic}
\end{minipage}
\par\smallskip
\textit{Contract:} all features are encoded once; query labels are never
read; $C_\theta$ and $H_\theta$ each execute once. Rejection sets
$\alpha=0$ and returns the exact Core representation.
\end{algorithm}

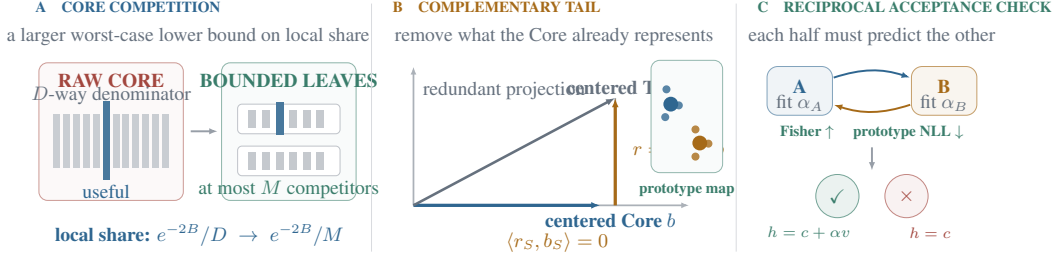
\begin{figure}[t]
\centering
\resizebox{0.94\linewidth}{!}{
\begin{tikzpicture}[x=1cm,y=1cm,
 note/.style={font=\small,text=scffgray,align=center},
 smallnote/.style={font=\footnotesize,text=scffgray,align=center},
 flow/.style={-{Latex[length=1.4mm]},line width=0.9pt,draw=scffgray!68}]
\path[use as bounding box] (0,0) rectangle (16,4.55);

\node[note,font=\scriptsize\bfseries,anchor=west,text=scffblue] at (0.10,4.22)
 {A\quad CORE COMPETITION};
\node[smallnote] at (2.70,3.73) {a larger worst-case lower bound on local share};

\draw[rounded corners=4pt,draw=scffred!55,fill=scffred!3]
 (0.28,1.08) rectangle (2.62,3.31);
\node[smallnote,text=scffred,font=\footnotesize\bfseries] at (1.45,3.03)
 {RAW CORE};
\foreach \i in {0,...,11}{
 \pgfmathsetmacro{\xx}{0.52+0.16*\i}
 \fill[scffgray!35] (\xx,1.61) rectangle ++(0.10,0.87);
}
\fill[scffblue!88] (1.32,1.41) rectangle ++(0.10,1.27);
\node[smallnote,text=scffblue] at (1.37,1.22) {useful};
\node[smallnote] at (1.45,2.78) {$D$-way denominator};

\draw[flow] (2.74,2.20)--(3.18,2.20);
\draw[rounded corners=4pt,draw=scffgreen!58,fill=scffgreen!3]
 (3.24,1.08) rectangle (5.42,3.31);
\node[smallnote,text=scffgreen,font=\footnotesize\bfseries] at (4.33,3.03)
 {BOUNDED LEAVES};
\foreach \y in {2.40,1.72}{
 \draw[rounded corners=2pt,draw=scffgray!42,fill=white]
 (3.50,\y-0.24) rectangle (5.16,\y+0.24);
 \foreach \i in {0,...,5}{
   \fill[scffgray!38] (3.68+0.22*\i,\y-0.14) rectangle ++(0.12,0.28);
 }
}
\fill[scffblue!88] (4.12,2.14) rectangle ++(0.12,0.52);
\node[smallnote,text=scffgreen] at (4.33,1.26) {at most $M$ competitors};
\node[smallnote,text=scffblue,font=\footnotesize\bfseries] at (2.88,0.52)
 {local share: $e^{-2B}/D\;\rightarrow\;e^{-2B}/M$};

\draw[scffgray!25] (5.67,0.25)--(5.67,4.27);

\node[note,font=\scriptsize\bfseries,anchor=west,text=scffamber] at (5.89,4.22)
 {B\quad COMPLEMENTARY TAIL};
\node[smallnote] at (8.63,3.73) {remove what the Core already represents};

\draw[-{Latex[length=1.3mm]},draw=scffgray!55,line width=0.8pt]
 (6.35,1.00)--(10.82,1.00);
\draw[-{Latex[length=1.3mm]},draw=scffgray!55,line width=0.8pt]
 (6.35,1.00)--(6.35,3.25);
\draw[-{Latex[length=1.5mm]},draw=scffblue,line width=1.5pt]
 (6.35,1.00)--(9.38,1.00);
\node[smallnote,text=scffblue,font=\footnotesize\bfseries] at (9.54,0.75) {centered Core $b$};
\draw[-{Latex[length=1.5mm]},draw=scffgray,line width=1.2pt]
 (6.35,1.00)--(9.62,2.74);
\node[smallnote,text=scffgray,font=\footnotesize\bfseries] at (9.84,2.91) {centered Tail $u$};
\draw[dashed,draw=scffgray!60,line width=0.8pt]
 (9.62,2.74)--(9.62,1.00);
\draw[-{Latex[length=1.5mm]},draw=scffamber,line width=1.5pt]
 (9.62,1.00)--(9.62,2.74);
\node[smallnote,text=scffamber,font=\footnotesize\bfseries,anchor=west]
 at (9.78,1.90) {$r=u-\rho b$};
\node[smallnote] at (7.82,2.86) {redundant projection};
\node[smallnote,text=scffamber] at (8.68,0.40) {$\langle r_S,b_S\rangle=0$};

\draw[rounded corners=4pt,draw=scffgreen!55,fill=scffgreen!3]
 (10.20,1.53) rectangle (11.36,3.35);
\foreach \x/\y in {10.45/2.88,10.65/2.62,10.43/2.43}{
 \fill[scffblue!80] (\x,\y) circle (0.065);
}
\foreach \x/\y in {10.92/2.23,11.13/2.00,10.92/1.79}{
 \fill[scffamber!80] (\x,\y) circle (0.065);
}
\fill[scffblue] (10.52,2.64) circle (0.12);
\fill[scffamber] (10.99,2.01) circle (0.12);
\node[smallnote,text=scffgreen,font=\scriptsize\bfseries] at (10.78,1.26)
 {prototype map};

\draw[scffgray!25] (11.58,0.25)--(11.58,4.27);

\node[note,font=\scriptsize\bfseries,anchor=west,text=scffgreen] at (11.80,4.22)
 {C\quad RECIPROCAL ACCEPTANCE CHECK};
\node[smallnote] at (13.78,3.73) {each half must predict the other};

\draw[rounded corners=5pt,draw=scffblue!60,fill=scffblue!7]
 (12.08,2.46) rectangle (13.14,3.25);
\node[note,text=scffblue,font=\small\bfseries] at (12.61,2.93) {A};
\node[smallnote] at (12.61,2.66) {fit $\alpha_A$};
\draw[rounded corners=5pt,draw=scffamber!65,fill=scffamber!7]
 (14.42,2.46) rectangle (15.48,3.25);
\node[note,text=scffamber,font=\small\bfseries] at (14.95,2.93) {B};
\node[smallnote] at (14.95,2.66) {fit $\alpha_B$};
\draw[flow,scffblue,bend left=18] (13.16,3.07) to (14.40,3.07);
\draw[flow,scffamber,bend left=18] (14.40,2.63) to (13.16,2.63);
\node[smallnote,text=scffgreen,font=\scriptsize\bfseries] at (13.78,2.18)
 {Fisher $\uparrow$ \quad prototype NLL $\downarrow$};

\draw[flow] (13.78,1.94)--(13.78,1.56);
\node[circle,draw=scffgreen!70,fill=scffgreen!10,minimum size=7mm,
 text=scffgreen,font=\small\bfseries] at (13.22,1.15) {$\checkmark$};
\node[circle,draw=scffred!65,fill=scffred!7,minimum size=7mm,
 text=scffred,font=\small\bfseries] at (14.34,1.15) {$\times$};
\node[smallnote,text=scffgreen,font=\scriptsize\bfseries] at (12.76,0.55)
 {$h=c+\alpha v$};
\node[smallnote,text=scffred,font=\scriptsize\bfseries] at (14.72,0.55)
 {$h=c$};

\end{tikzpicture}}
\caption{\textbf{Why bounded folding can improve prediction.}
Bounded leaves increase local share; the optional Tail path removes Core-aligned
content and accepts only support-consistent residual directions.}
\label{fig:information-compression}
\end{figure}

\subsection{Combine strong evidence, test what the Tail adds}
\label{sec:tail-operator}

The Core provides the initial representation. SCFF averages its leaf messages into
$c=|\mathcal G_C|^{-1}\sum_{g\in\mathcal G_C}m_g$. This lets the feature
encoder model rich interactions among at most $M$ strong features at a time,
then combines the resulting row representations at the model's standard interface.

The remaining question is whether the Tail adds information beyond this
representation. SCFF averages its leaf messages, removes the global
direction already aligned with the centered Core representation, and considers both
the resulting raw direction and a support-prototype projection. The prototype
map is support-fitted: query rows use all-support class centroids, while each
support row uses a leave-one-out own-class centroid. A class-balanced split
then fits a Fisher step on each half and tests that step on the opposite half;
reciprocal nearest-prototype checks evaluate the fixed candidate in both
directions. Because the candidate prototype uses the full support set before
these checks, this is an acceptance test, not independent cross-fitting or a
separate validation set. The candidate must improve both checks; otherwise
$\alpha=0$ and SCFF returns the exact Core representation. Multiclass episodes also use
this fallback. The resulting representation
\begin{equation}
  h=c+\alpha v,\qquad
  p(y_q\mid X_s,y_s,X_q)=H_\theta(C_\theta(h;y_s))
  \label{eq:predict-once}
\end{equation}
produces one complete query distribution. Prototype scores test a candidate
direction; they are never averaged model predictions. Appendix
\ref{app:scff-proof} gives the projection, prototype map, reciprocal support-subset test,
and full derivations.

\subsection{Why the change can help}
\label{sec:budget-risk}

\paragraph{Bounded competition changes the inductive bias.}
If one feature-attention normalization has logits in $[-B,B]$, a feature among
$K$ competitors receives weight at least $e^{-2B}/K$. Full-width execution
therefore gives a bound of $e^{-2B}/D$, while a leaf of width at most $M$
gives $e^{-2B}/M$. The lower bound on local share increases by $D/M$ before
leaf aggregation. This motivates a predictive hypothesis: bounded groups can
give useful features more local influence. Aggregation can attenuate or cancel
that change, so matched Raw-Core and Bounded-Core controls measure its net effect.

\paragraph{Additional evidence must earn its place.}
After projection, the raw Tail residual is orthogonal to the centered Core on
the support rows. Reciprocal subset checks then require its class evidence to transfer in both
directions. If that evidence is absent, the update is
zero and the combined representation is exactly the Core representation.
The held-out Tail controls test whether support-accepted evidence also improves unseen
queries; Appendix~\ref{app:scff-proof} states the corresponding transfer
condition.

\paragraph{The expensive feature interaction becomes width-linear.}
For a pairwise feature mixer, a width-$D$ feature encoder costs
$\Theta(TD^2d)$. With $G\simeq D/M$ bounded leaves, the total becomes
\begin{equation}
  G\,\Theta(TM^2d)=\Theta(TDMd),
  \label{eq:feature-complexity}
\end{equation}
while sequential execution keeps an $O(TM^2)$ feature-interaction working set.
The $G$ representations are combined into one $T\times d_{\rm ICL}$ representation,
so the context encoder and label head still run once at the original context length.

\subsection{Reusing the same idea across architectures}
\label{sec:architecture-interface}

SCFF requires compatible row representations before a shared context encoder
and must reproduce native inference. TabPFN-3, TabICLv1/v2, TabPFNv1, and
RefineICL-24L~\citep{refineicl2026companion} expose
Equation~\eqref{eq:model-interface}. LimiX keeps a
feature-token axis; its adapter retains Core slots and folds the remainder into
bounded Tail slots with support-RMS rescaling (Appendix~\ref{app:limix-token-adapter}).

\section{Evaluating the Evidence--Memory Tradeoff}
\label{sec:experiments}

The empirical question is whether making a wide table cheaper to encode also
makes its evidence more useful. We first compare SCFF with native inference,
then hold the selected features fixed to isolate folding, and finally let
the memory savings pay for additional evidence. These comparisons distinguish
the benefit of choosing features from the benefit of processing them differently.

\paragraph{Benchmark: 18 wide-table datasets.}
Our main evaluation uses \textbf{18 classification datasets from AMLB-29,
the released TabZilla collection, and TabArena v0.1}, spanning 128--7,200 raw
features. These comprise all eligible dataset versions in the fixed registry
snapshot: at least 128 features, 2--50 classes, no missing target, and no
semantic duplicate. Registry metadata fixes this population before any model
outcome is read. We use fold 0 of these datasets for the explicitly described
finite method selection and reserve folds 1--4 for locked confirmation.
Confirmation uses new query rows on the same datasets.
The dataset list and benchmark membership appear in Appendix
Table~\ref{tab:public-wide-datasets}.

\paragraph{Models and evaluation.}
We evaluate five released backbones and RefineICL-24L, a frozen checkpoint
from a concurrent anonymous submission~\citep{refineicl2026companion}, over
five fixed folds.
Each native--SCFF pair uses the same checkpoint, preprocessing, split, and
query rows. We average folds within datasets, then macro-average datasets,
reporting paired point estimates for accuracy, NLL, time, and peak GPU memory.
Paired dataset-bootstrap intervals use these within-dataset averages.
Four backbones have matched-width comparisons; LimiX tests a token-aligned
adapter, and TabPFNv1 tests a capacity extension. Coverage and adapter
contracts appear separately in Appendix Table~\ref{tab:evidence-tiers}.
We use open-source benchmark data and model checkpoints, and provide our
reproduction code in the supplementary material.

\paragraph{Freezing the design before confirmation.}
Fold 0 selects the parameter-free Tail update, prioritizing the number of
compatible backbones improving both metrics, then mean NLL and accuracy.
We lock the method and configuration before folds 1--4. The complete-method
confirmation and the incremental Tail contrast remain separate: the latter
requires at least three of five compatible backbones to improve both metrics.
All five are reported in Appendix Table~\ref{tab:tail-confirmation-complete};
the complete-method intervals appear in Table~\ref{tab:heldout-full-method}.
Deterministic stratified query folds are disjoint and support sets are their
complements. Seeds, caps, checkpoints, and execution details appear in
Appendix Table~\ref{tab:reproduction-contract}.

\paragraph{Separating selection, folding, and recovery.}
\emph{Raw Core} sends the ranked Core through one standard feature-encoder call;
\emph{Bounded Core} folds the same columns into leaves of at most $M=128$.
\emph{Tail Off} sets $\alpha=0$; \emph{Tail On} adds the selected Tail update.
These pairs ask whether folding improves an already selected feature set
and whether the Tail contributes beyond it. Tail On encodes every feature
once; rejection reproduces Tail Off.

\paragraph{Testing the design choices.}
Table~\ref{tab:full-coverage-component-ablation} tests seven fixed TabICLv2
alternatives on the same 18 datasets and five folds: random routing, no Core
projection, and a sequence of Tail variants that adds the residual, prototype
map, Fisher step, reciprocal support-subset acceptance, single-leaf fallback, and binary-task
restriction. Every variant retains all features in one prediction; deltas are
full SCFF minus each complete variant, not additive effects. Random routing preserves leaf sizes and
coverage, while the projection control changes only the residual geometry.

\paragraph{Turning memory savings into evidence.}
We compare Bounded Core with the widest support-ranked single leaf fitting
its measured peak-memory budget; memory alone selects the prefix before
quality is compared. Separate interventions vary leaf width, add up to 512
label-independent columns, and increase nested support context on fixed query
rows. An earlier all-feature mass-Huber variant additionally tests context
reinvestment under Native@2,048's measured budget (Appendix~\ref{app:experimental-detail}).

\section{Results}
\label{sec:results}

\subsection{Wide-table quality improves with a smaller memory footprint}

\begin{center}\begin{minipage}{\linewidth}
\centering
\captionof{table}{\textbf{Main results on the fixed 18-dataset wide-table registry from AMLB-29, TabZilla, and TabArena.} The panel spans 128--7,200 raw features. We compare SCFF with native inference over five fixed folds within each backbone's native evaluation contract.}
\label{tab:full-coverage-main}
\footnotesize
\setlength{\tabcolsep}{2.5pt}
\begin{tabular}{@{}lrrrrrr@{}}
\toprule
Backbone & $\Delta$Acc. pp $\uparrow$ & Error $\downarrow$ & $\Delta$NLL $\downarrow$ & NLL $\downarrow$ & Median E2E $\uparrow$ & Median mem. $\downarrow$ \\
\midrule
TabPFN-3 & +1.874 & +11.8\% & -0.05065 & +11.8\% & 1.00$\times$ & 2.15$\times$ \\
TabICLv1 & +5.419 & +26.1\% & -0.09099 & +18.3\% & 0.62$\times$ & 2.36$\times$ \\
TabICLv2 & +1.564 & +11.1\% & -0.03557 & +9.6\% & 0.56$\times$ & 2.32$\times$ \\
TabPFNv1$^\ddagger$ & +3.071 & +13.9\% & -0.04341 & +8.5\% & -- & -- \\
LimiX-2M$^\dagger$ & +0.700 & +4.7\% & -0.01178 & +3.2\% & 1.72$\times$ & 2.46$\times$ \\
RefineICL-24L & +3.730 & +22.0\% & -0.08223 & +19.0\% & 4.84$\times$ & 2.09$\times$ \\
\bottomrule
\end{tabular}
\vspace{2pt}

\parbox{0.98\linewidth}{\footnotesize Quality is dataset-macro; resources are native/SCFF medians over verified paired folds and E2E includes support compilation. Error and NLL are relative reductions. Model-specific native eligibility, $\ddagger$capacity and $\dagger$token contracts, and full resource distributions are detailed in SI, Appendix Tables~\ref{tab:evidence-tiers} and~\ref{tab:paired-resource-distribution}.}
\end{minipage}\end{center}

The first result is that the memory reduction need not come at the expense
of predictive quality. SCFF improves accuracy and NLL on all six evaluated
backbones. On the dataset-macro aggregate, relative error falls by up to $26.1\%$ and relative
NLL by up to $19.0\%$. All four matched-width rows
have favorable locked-fold dataset-bootstrap intervals for both metrics
(Appendix Table~\ref{tab:heldout-full-method}); the separately reported
capacity and token-interface extensions cross zero. Across the four
matched-width backbones,
fold-paired median memory savings are $2.09$--$2.36\times$; ratios of the
largest observed native and SCFF peaks reach $5.2$--$34.3\times$.
Median end-to-end speed is near
parity for TabPFN-3, slower for both TabICL backbones, and faster for LimiX
and RefineICL. Contracts and intervals appear in Appendix
Tables~\ref{tab:evidence-tiers} and~\ref{tab:headline-bootstrap}. Leave-one-dataset-out
effects remain favorable for all matched-width backbones
(Appendix Table~\ref{tab:heldout-influence}); paired resource distributions
appear in Appendix Table~\ref{tab:paired-resource-distribution}. The next
comparisons ask how much of this improvement comes from selection and how
much from the way the selected evidence is processed.

\subsection{Bounded Core and Tail recovery contribute distinct gains}

\begin{center}\begin{minipage}{\linewidth}
\centering
\captionof{table}{\textbf{Selection supplies quality; folding supplies capacity.} Panel A decomposes native $\rightarrow$ selected Core $\rightarrow$ bounded Core on 18 datasets and five folds; Panel B adds optional Tail evidence.}
\label{tab:core-tail-targets}
\footnotesize
\renewcommand{\arraystretch}{0.88}
\setlength{\tabcolsep}{4.0pt}
\begin{tabular}{@{}lrrr@{}}
\toprule
Backbone & \shortstack{Selection $-$ native\\$\Delta$Acc. pp / $\Delta$NLL} & \shortstack{Bounded $-$ selection\\$\Delta$Acc. pp / $\Delta$NLL} & Raw/Bounded max. peaks $\uparrow$ \\
\midrule
\multicolumn{4}{@{}l}{\textit{A. Full-panel Core decomposition (18 datasets, five folds)}} \\
TabPFN-3 & \shortstack{+1.563\\[-1pt]{\scriptsize -0.03414}} & \shortstack{+0.233\\[-1pt]{\scriptsize -0.00639}} & 30.1$\times$ \\
TabICLv1 & \shortstack{+5.175\\[-1pt]{\scriptsize -0.07959}} & \shortstack{+0.162\\[-1pt]{\scriptsize -0.00911}} & 32.3$\times$ \\
TabICLv2 & \shortstack{+0.690\\[-1pt]{\scriptsize -0.01576}} & \shortstack{+0.739\\[-1pt]{\scriptsize -0.01683}} & 34.0$\times$ \\
\addlinespace
\multicolumn{4}{@{}l}{\textit{B. Optional Tail extraction: Tail On $-$ Tail Off (held-out folds 1--4)}} \\
TabPFN-3 & \shortstack{+0.107\\[-1pt]{\scriptsize -0.00985}} & -- & -- \\
TabICLv1 & \shortstack{+0.080\\[-1pt]{\scriptsize -0.00234}} & -- & -- \\
TabICLv2 & \shortstack{+0.145\\[-1pt]{\scriptsize -0.00294}} & -- & -- \\
\bottomrule
\end{tabular}
\vspace{2pt}

\parbox{0.98\linewidth}{\footnotesize Cells report $\Delta$Accuracy pp / $\Delta$NLL. Selection supplies most quality; folding adds a smaller gain on all three backbones. The 30.1--34.0$\times$ ratios divide the separate largest raw and bounded Core peaks. Appendix Tables~\ref{tab:core-confirmation-intervals} and~\ref{tab:tail-confirmation-complete} report intervals and all Tail-compatible models.}
\end{minipage}\end{center}

Feature selection alone already helps. Across the three backbones, it identifies a strong Core
($+0.690$ to $+5.175$ accuracy points), and folding that same Core improves
both predictive metrics again. Thus the selected set does not fully explain
the gains. The corresponding ratio between the largest
observed raw and bounded Core peaks is $30.1$--$34.0\times$. Both
all-five-fold folding intervals exclude zero on TabICLv2. A secondary
folds 1--4 same-Core analysis also finds both intervals favorable on
TabPFN-3; TabICLv1 remains uncertain
(Appendix Table~\ref{tab:heldout-core-quality}).
The Core need not be the end of the evidence budget. On held-out folds, Tail On improves both
point estimates on TabPFN-3, TabICLv1, and TabICLv2 while encoding every
remaining column and retaining exact Core fallback. Appendix
Tables~\ref{tab:core-confirmation-intervals} and
\ref{tab:tail-confirmation-complete} give the full intervals and all five
compatible models.

\subsection{How evidence is organized matters}

Splitting a table is not sufficient by itself. The following controls keep
coverage and prediction count fixed while changing which features share a
role and how additional evidence enters the representation.

\begin{center}
\begin{minipage}{\linewidth}
\centering
\captionof{table}{\textbf{Component controls around full SCFF.} Each row reports a paired TabICLv2 point estimate, full SCFF minus one fixed alternative; positive accuracy and negative NLL favor SCFF.}
\label{tab:full-coverage-component-ablation}
\footnotesize
\setlength{\tabcolsep}{6.0pt}
\begin{tabular}{@{}lrr@{}}
\toprule
Ablated variant & $\Delta$Acc. pp $\uparrow$ & $\Delta$NLL $\downarrow$ \\
\midrule
Random Core/Tail assignment & $+1.878$ & $-0.03811$ \\
Centered Tail (no Core projection) & $+0.054$ & $-0.00125$ \\
Raw residual only & $+0.018$ & $-0.00403$ \\
Fixed prototype update & $+0.586$ & $-0.01680$ \\
Fisher-selected update & $+0.513$ & $-0.01481$ \\
Validated update, no single-leaf fallback & $+0.209$ & $-0.00874$ \\
Validated update, no binary-task restriction & $+0.202$ & $-0.00881$ \\
\bottomrule
\end{tabular}
\vspace{2pt}
\parbox{0.98\linewidth}{\footnotesize 18 datasets $\times$ five folds; every variant retains all features and produces one prediction. The first two change one component; the last five build the Tail update step by step. Table~\ref{tab:core-tail-targets} compares the complete Core and Tail paths across backbones.}
\end{minipage}
\end{center}

\textbf{Core allocation.}
Random routing preserves leaf sizes, coverage, and inference count, yet falls
below native by $0.314$ accuracy points and $0.00254$ NLL; full SCFF is $1.878$
points and $0.03811$ NLL better. The grouping must therefore reflect the
task's support evidence. Removing Core projection costs $0.054$ points and
$0.00125$ NLL, consistent with the value of separating new from redundant evidence.

\textbf{Tail acceptance.}
Raw residual nearly matches accuracy but worsens NLL by $0.00403$;
fixed-prototype and Fisher-selected updates trail SCFF by $0.586/0.01680$ and
$0.513/0.01481$ accuracy points/NLL. Reciprocal support-subset checking gives
the best point estimates among these fixed controls.

\textbf{Keeping the evidence fixed.}
The same-$K$ comparison provides the resource counterpart to the Core quality
control. Across 30 paired units, bounded leaves use 13.4\% of the single-leaf
incremental peak memory at $1.41\times$ latency. The same selected evidence
has become substantially cheaper in memory. Section~\ref{sec:memory-payoff}
tests whether spending this saving on additional features improves prediction.

\subsection{Moderate leaf width preserves useful interaction capacity}

\begin{center}
\begin{minipage}{\linewidth}
\centering
\captionof{table}{\textbf{Moderate leaves preserve interactions.} All columns retained; four datasets, five folds.}
\label{tab:full-coverage-leaf-width}
\footnotesize
\setlength{\tabcolsep}{5.0pt}
\begin{tabular}{@{}rrrrrr@{}}
\toprule
$M$ & Median leaves & $\Delta$Acc. pp $\uparrow$ & $\Delta$NLL $\downarrow$ & Median time (s) & Peak GPU (GiB) \\
\midrule
32 & 40.5 & $-0.498$ & $+0.00872$ & 3.135 & 1.986 \\
64 & 21.0 & $+0.287$ & $-0.01310$ & 2.666 & 1.100 \\
\textbf{128} & 11.0 & $\mathbf{+0.255}$ & $\mathbf{-0.01070}$ & 3.281 & 1.666 \\
\bottomrule
\end{tabular}
\vspace{2pt}

\parbox{0.98\linewidth}{\footnotesize Deltas are versus native; peak is the maximum over 20 runs. Fold 0 selected $M=128$.}
\end{minipage}
\end{center}

The leaf cap balances affordable interactions against the information lost
when related features are separated. $M=32$ loses quality, whereas 64 and
128 improve over native. The full evaluation retains the fold-0-selected
$M=128$. The benefit comes from bounded but expressive groups, rather than
making the groups as small as possible.

\subsection{Memory savings buy more useful evidence}
\label{sec:memory-payoff}

The practical test is what a model can do with the released memory budget.
We examine two uses: accommodating more examples and retaining a wider Core.

\begin{center}
\begin{minipage}{\linewidth}
\centering
\captionof{table}{\textbf{Width determines how SCFF responds to more context.}
Each cell is $\Delta$accuracy points / $\Delta$NLL relative to native inference
on the same nested support prefix and fixed query rows.}
\label{tab:full-coverage-context-width}
\footnotesize
\setlength{\tabcolsep}{6.0pt}
\begin{tabular}{@{}lrrrr@{}}
\toprule
Dataset & Raw $D$ & Context 128 & Context 512 & Context 2,048 \\
\midrule
\texttt{kddcup09\_appetency} & 212
  & $+0.000/-0.00791$ & $+0.000/+0.00408$ & $+0.000/-0.00166$ \\
\texttt{philippine} & 308
  & $-4.102/+0.08142$ & $-2.246/+0.00761$ & $+0.879/-0.01000$ \\
\texttt{dilbert} & 2,000
  & $+5.762/-0.18864$ & $+2.148/-0.01944$ & $-0.098/+0.00049$ \\
\texttt{guillermo} & 4,296
  & $+2.344/-0.08459$ & $+8.594/-0.10163$ & $+8.887/-0.11406$ \\
\bottomrule
\end{tabular}
\vspace{2pt}

\parbox{0.98\linewidth}{\footnotesize The four datasets are fixed from raw-width
bands before evaluation. Appendix Table~\ref{tab:full-coverage-context-width-detail}
reports memory, time, and the complete protocol for all 12 cells.}
\end{minipage}
\end{center}

At 4,296 features, the gain over native rises from $2.344$ to $8.887$ accuracy
points as both arms receive more context. The 212-feature task stays near
native, with mixed intermediate responses. Across all four, peak memory falls
from $37.6\%$ to $27.0\%$ of native as context grows from 128 to 2,048 rows.
This motivates a stricter comparison: require the single-leaf baseline to fit
within the folded route's measured memory ceiling.

On the predeclared $K>128$ strata, folding retains a median 244 versus 96
features on TabICLv2, gaining $4.06$ accuracy points and reducing NLL by
$0.100$ across 10 datasets. The same rule gains $3.72$ points and reduces NLL
by $0.086$ on nine TabPFN-3 datasets. Figure~\ref{fig:equal-memory-capacity-quality}
connects the extra retained evidence to these gains and their 95\% intervals.

\noindent\begin{minipage}{\linewidth}
\centering
\includegraphics[width=\linewidth]{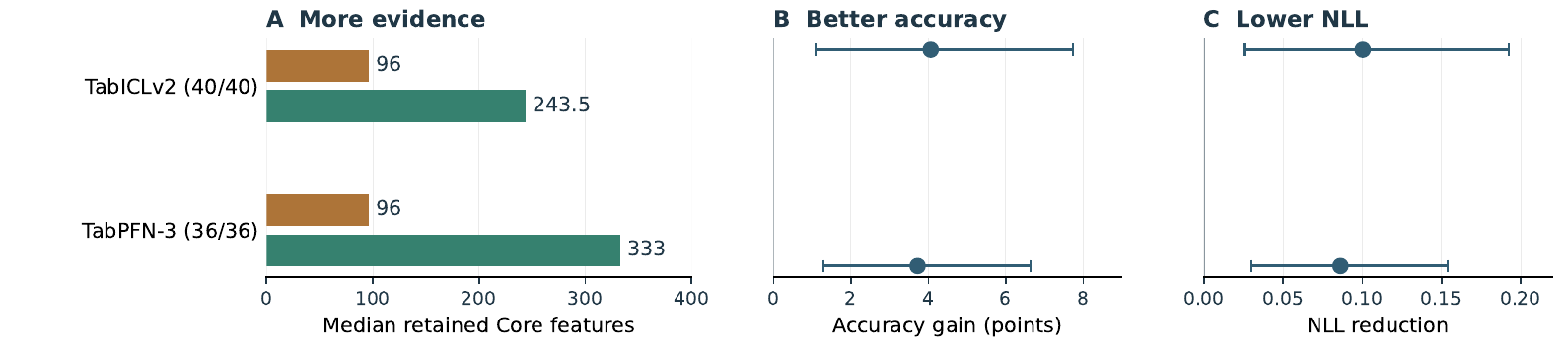}
\captionof{figure}{\textbf{More evidence below a measured peak-memory ceiling improves prediction.}
Predeclared $K>128$ datasets: TabICLv2 (10), TabPFN-3 (9), four folds each.
The widest memory-feasible prefix from a fixed grid (ochre) stays below the
paired folded peak (green) in all 40/40 and 36/36 folds; folded/prefix peak ratios are $1.17\times$
(median) and $1.13\times$ (mean). Accuracy/NLL effects show dataset-bootstrap
95\% intervals. Both routes make one full prediction.}
\label{fig:equal-memory-capacity-quality}
\end{minipage}\par

\begin{center}\begin{minipage}{\linewidth}
\centering
\captionof{table}{\textbf{Equal-memory gains extend beyond the highest-width strata.}
Each comparison uses one frozen model member and four held-out folds per dataset.
The single leaf is the widest support-ranked prefix below the paired folded peak
in the fixed width grid; intervals resample datasets.}
\label{tab:main-equal-memory-scope}
\small
\setlength{\tabcolsep}{5pt}
\begin{tabular}{@{}llrrll@{}}
\toprule
Backbone & Scope & Data & Folds & $\Delta$Acc. pp [95\% CI] & NLL reduction [95\% CI] \\
\midrule
TabICLv2 & Full registry$^\dagger$ & 18 & 72 & $+2.28\ [ +0.50,+4.63]$ & $0.0560\ [0.0112,0.1149]$ \\
 & Strict complete cases & 16 & 64 & $+2.56\ [ +0.60,+5.16]$ & $0.0630\ [0.0134,0.1288]$ \\
 & Core width $K>128$ & 10 & 40 & $+4.06\ [ +1.09,+7.73]$ & $0.1002\ [0.0252,0.1926]$ \\
TabPFN-3 & Full registry & 18 & 72 & $+1.86\ [ +0.47,+3.63]$ & $0.0431\ [0.0110,0.0841]$ \\
 & Core width $K>128$ & 9 & 36 & $+3.72\ [ +1.29,+6.63]$ & $0.0862\ [0.0302,0.1541]$ \\
\bottomrule
\end{tabular}
\vspace{2pt}

\parbox{0.98\linewidth}{\footnotesize $^\dagger$Six TabICLv2 one-leaf identity folds have equal predictions but a $0.19$--$1.39\%$ difference in one-shot peak readings. Full-registry includes these zero-effect pairs; strict complete cases retain the 16 datasets with all four folds within budget. Per-dataset results and protocols appear in Appendix Tables~\ref{tab:equal-memory-confirmation} and~\ref{tab:equal-memory-tabpfn3}.}
\end{minipage}\end{center}

Both backbones also improve across the complete registry. Since memory alone
fixes each prefix before quality is measured, this comparison completes the
argument: saved memory buys more evidence for the same frozen predictor.

\paragraph{Conclusion.}
Using more features need not require interacting over all of them at once.
SCFF improves six frozen backbones through bounded feature computations and
one contextual prediction. Its memory savings can preserve more useful
evidence, turning width adaptation into an inference-time opportunity for
tabular foundation models.

\label{marker:main-text-end}

\clearpage
\bibliography{references}

\begin{thebibliography}{19}
\providecommand{\natexlab}[1]{#1}
\providecommand{\url}[1]{\texttt{#1}}
\expandafter\ifx\csname urlstyle\endcsname\relax
  \providecommand{\doi}[1]{doi: #1}\else
  \providecommand{\doi}{doi: \begingroup \urlstyle{rm}\Url}\fi

\bibitem[{Anonymous}(2026)]{refineicl2026companion}
{Anonymous}.
\newblock What does tabular icl compute? in-situ representation refinement
  through attention-gated updates, 2026.
\newblock Concurrent submission to ICLR 2027.

\bibitem[Benjamini \& Hochberg(1995)Benjamini and
  Hochberg]{benjamini1995controlling}
Yoav Benjamini and Yosef Hochberg.
\newblock Controlling the false discovery rate: A practical and powerful
  approach to multiple testing.
\newblock \emph{Journal of the Royal Statistical Society: Series B},
  57\penalty0 (1):\penalty0 289--300, 1995.
\newblock \doi{10.1111/j.2517-6161.1995.tb02031.x}.
\newblock URL \url{https://academic.oup.com/jrsssb/article/57/1/289/7035855}.

\bibitem[Eo et~al.(2026)Eo, Suh, Cho, Kim, Kim, Nam, and Lee]{eo2026exaone}
Moonjung Eo, Min-Kook Suh, Hye-Seung Cho, Jiwon Kim, Seoyoon Kim, Sangjun Nam,
  and Soonyoung Lee.
\newblock {EXAONE Tabular 1.0}: Technical report.
\newblock \emph{arXiv preprint arXiv:2608.25774}, 2026.
\newblock URL \url{https://arxiv.org/abs/2608.25774}.

\bibitem[{Google Research}(2026)]{tabfmcode2026}
{Google Research}.
\newblock {TabFM}: Tabular foundation models.
\newblock Software repository, 2026.
\newblock URL \url{https://github.com/google-research/tabfm}.
\newblock Repository accessed September 18, 2026; no technical report was
  available.

\bibitem[Grinsztajn et~al.(2025)Grinsztajn, Fl{\"o}ge, Key,
  et~al.]{grinsztajn2025tabpfn25}
L{\'e}o Grinsztajn, Klemens Fl{\"o}ge, Oscar Key, et~al.
\newblock {TabPFN-2.5}: Advancing the state of the art in tabular foundation
  models.
\newblock \emph{arXiv preprint arXiv:2511.08667}, 2025.
\newblock URL \url{https://arxiv.org/abs/2511.08667}.

\bibitem[Grinsztajn et~al.(2026)Grinsztajn, Fl{\"o}ge, Key,
  et~al.]{grinsztajn2026tabpfn3}
L{\'e}o Grinsztajn, Klemens Fl{\"o}ge, Oscar Key, et~al.
\newblock {TabPFN-3}: Technical report.
\newblock \emph{arXiv preprint arXiv:2605.13986}, 2026.
\newblock URL \url{https://arxiv.org/abs/2605.13986}.

\bibitem[Guyon \& Elisseeff(2003)Guyon and Elisseeff]{guyon2003variable}
Isabelle Guyon and Andr{\'e} Elisseeff.
\newblock An introduction to variable and feature selection.
\newblock \emph{Journal of Machine Learning Research}, 3:\penalty0 1157--1182,
  2003.
\newblock URL \url{https://www.jmlr.org/papers/v3/guyon03a.html}.

\bibitem[Hollmann et~al.(2025)Hollmann, M{\"u}ller, Purucker,
  et~al.]{hollmann2025tabpfn}
Noah Hollmann, Samuel M{\"u}ller, Lennart Purucker, et~al.
\newblock Accurate predictions on small data with a tabular foundation model.
\newblock \emph{Nature}, 637:\penalty0 319--326, 2025.
\newblock \doi{10.1038/s41586-024-08328-6}.
\newblock URL \url{https://doi.org/10.1038/s41586-024-08328-6}.

\bibitem[Kolberg et~al.(2025)Kolberg, Kreuer, Huurdeman, Ouaari, Eggensperger,
  and Pfeifer]{kolberg2025wide}
Christopher Kolberg, Jules Kreuer, Jonas Huurdeman, Sofiane Ouaari, Katharina
  Eggensperger, and Nico Pfeifer.
\newblock {TabPFN-Wide}: Continued pre-training for extreme feature counts.
\newblock \emph{arXiv preprint arXiv:2510.06162}, 2025.
\newblock URL \url{https://arxiv.org/abs/2510.06162}.

\bibitem[Lee et~al.(2019)Lee, Lee, Kim, Kosiorek, Choi, and Teh]{lee2019set}
Juho Lee, Yoonho Lee, Jungtaek Kim, Adam Kosiorek, Seungjin Choi, and Yee~Whye
  Teh.
\newblock Set transformer: A framework for attention-based
  permutation-invariant neural networks.
\newblock In \emph{Proceedings of the 36th International Conference on Machine
  Learning}, volume~97 of \emph{Proceedings of Machine Learning Research}, pp.\
   3744--3753, 2019.
\newblock URL \url{https://proceedings.mlr.press/v97/lee19d.html}.

\bibitem[M{\"u}ller et~al.(2022)M{\"u}ller, Hollmann, Pineda~Arango, Grabocka,
  and Hutter]{mueller2022pfn}
Samuel M{\"u}ller, Noah Hollmann, Sebastian Pineda~Arango, Josif Grabocka, and
  Frank Hutter.
\newblock Transformers can do bayesian inference.
\newblock In \emph{International Conference on Learning Representations}, 2022.
\newblock URL \url{https://openreview.net/forum?id=u3vyZf5Jv2}.

\bibitem[Qu et~al.(2025)Qu, Holzm{\"u}ller, Varoquaux, and
  Le~Morvan]{qu2025tabicl}
Jingang Qu, David Holzm{\"u}ller, Ga{\"e}l Varoquaux, and Marine Le~Morvan.
\newblock {TabICL}: A tabular foundation model for in-context learning on large
  data.
\newblock In \emph{Proceedings of the 42nd International Conference on Machine
  Learning}, volume 267 of \emph{Proceedings of Machine Learning Research},
  pp.\  50817--50847, 2025.
\newblock URL \url{https://proceedings.mlr.press/v267/qu25d.html}.

\bibitem[Qu et~al.(2026)Qu, Holzm{\"u}ller, Varoquaux, and
  Le~Morvan]{qu2026tabiclv2}
Jingang Qu, David Holzm{\"u}ller, Ga{\"e}l Varoquaux, and Marine Le~Morvan.
\newblock {TabICLv2}: A better, faster, scalable, and open tabular foundation
  model.
\newblock \emph{arXiv preprint arXiv:2602.11139}, 2026.
\newblock URL \url{https://arxiv.org/abs/2602.11139}.

\bibitem[Silla \& Freitas(2011)Silla and Freitas]{silla2011hierarchical}
Carlos~N. Silla, Jr. and Alex~A. Freitas.
\newblock A survey of hierarchical classification across different application
  domains.
\newblock \emph{Data Mining and Knowledge Discovery}, 22\penalty0
  (1--2):\penalty0 31--72, 2011.
\newblock \doi{10.1007/s10618-010-0175-9}.
\newblock URL \url{https://doi.org/10.1007/s10618-010-0175-9}.

\bibitem[Thomas et~al.(2024)Thomas, Ma, Hosseinzadeh, Golestan, Yu, Volkovs,
  and Caterini]{thomas2024retrieval}
Valentin Thomas, Junwei Ma, Rasa Hosseinzadeh, Keyvan Golestan, Guangwei Yu,
  Maksims Volkovs, and Anthony Caterini.
\newblock Retrieval \& fine-tuning for in-context tabular models.
\newblock In \emph{Advances in Neural Information Processing Systems},
  volume~37, 2024.
\newblock URL
  \url{https://proceedings.neurips.cc/paper_files/paper/2024/hash/c40daf14d7a6469e65116507c21faeb7-Abstract-Conference.html}.

\bibitem[Tschalzev et~al.(2026)Tschalzev, Erickson, Wang, Rangwala, L{\"u}dtke,
  Stuckenschmidt, and Bartelt]{tschalzev2026tabprep}
Andrej Tschalzev, Nick Erickson, Yuyang Wang, Huzefa Rangwala, Stefan
  L{\"u}dtke, Heiner Stuckenschmidt, and Christian Bartelt.
\newblock {TabPrep}: Closing the feature engineering gap in tabular benchmarks.
\newblock \emph{arXiv preprint arXiv:2606.02384}, 2026.
\newblock URL \url{https://arxiv.org/abs/2606.02384}.

\bibitem[Ye et~al.(2025)Ye, Liu, and Chao]{ye2025closer}
Han-Jia Ye, Si-Yang Liu, and Wei-Lun Chao.
\newblock A closer look at {TabPFN} v2: Understanding its strengths and
  extending its capabilities.
\newblock \emph{arXiv preprint arXiv:2502.17361}, 2025.
\newblock URL \url{https://arxiv.org/abs/2502.17361}.

\bibitem[Zaheer et~al.(2017)Zaheer, Kottur, Ravanbakhsh, P{\'o}czos,
  Salakhutdinov, and Smola]{zaheer2017deep}
Manzil Zaheer, Satwik Kottur, Siamak Ravanbakhsh, Barnab{\'a}s P{\'o}czos,
  Ruslan Salakhutdinov, and Alexander~J. Smola.
\newblock Deep sets.
\newblock In \emph{Advances in Neural Information Processing Systems},
  volume~30, 2017.
\newblock URL
  \url{https://proceedings.neurips.cc/paper/2017/hash/f22e4747da1aa27e363d86d40ff442fe-Abstract.html}.

\bibitem[Zhang et~al.(2025)Zhang, Ren, Yu, Yuan, Wang, Li, et~al.]{limix2025}
Xingxuan Zhang, Gang Ren, Han Yu, Hao Yuan, Hui Wang, Jiansheng Li, et~al.
\newblock {LimiX}: Unleashing structured-data modeling capability for
  generalist intelligence.
\newblock \emph{arXiv preprint arXiv:2509.03505}, 2025.
\newblock URL \url{https://arxiv.org/abs/2509.03505}.

\end{thebibliography}
\bibliographystyle{iclr2027_conference}

\subsection*{Reproducibility statement}
The primary matched-fold comparisons are indexed by checkpoint hashes, data
manifests, fold records, adapter revisions, and matched native--SCFF runs.
Appendix~\ref{app:protocol} specifies the inference and statistical contracts.
The supplementary archive includes code, manifests, paired fold-level metric
records and table renderers for the primary locked-fold evaluation, and its
failure histories. It does not include per-query predictions or model weights.

\subsection*{AI use statement}
Generative AI tools were used to assist with code implementation, debugging,
experiment orchestration, and manuscript editing. The authors defined the
research questions and protocols, reviewed all generated code and prose, and
verified reported results against executable tests and hash-audited artifacts.

\ifdefined\SCFFMainOnly\else
\appendix
\section{Formal Properties and Scope}
\label{app:scff-proof}

The central design claim is that a wide input can be processed through bounded
feature interactions and still supply one contextual prediction. We first
establish how this changes computation, then explain how the Tail is tested
for information beyond the Core. This connects the resource argument to the
predictive hypothesis: coverage and bounded interactions follow from the
construction, while better predictions depend on the frozen representation
and the transfer of support evidence to unseen rows.

\subsection{Feature coverage and resource scaling}

Folding would be a trivial memory reduction if it simply dropped columns.
The first property rules out that explanation: the computational units get
narrower while the encoded feature set remains complete.

Within Core and Tail, SCFF stably ranks columns, distributes them round-robin
over the minimum number of width-$M$ leaves, and restores original column order
inside each leaf. For the resulting leaves $P_1,\ldots,P_G$,
\begin{equation}
 P_g\cap P_{g'}=\varnothing\;(g\ne g'),\qquad
 \bigcup_{g=1}^{G}P_g=[D],\qquad 1\le |P_g|\le M.
 \label{eq:full-partition}
\end{equation}
Equation~\eqref{eq:full-partition} follows because the two
strata are disjoint and exhaustive and the distribution changes neither
membership nor multiplicity. Consequently, every preprocessed feature is
encoded exactly once. This guarantees feature coverage; it does not reconstruct the nonlinear
full-width representation.

This changes the competition faced by each feature before the row
representations are combined. Suppose one feature-attention row has logits
$a_1,\ldots,a_K\in[-B,B]$. For any index $j$,
\begin{equation}
 \operatorname{softmax}(a)_j
 =\frac{e^{a_j}}{\sum_{k=1}^K e^{a_k}}
 \ge \frac{e^{-B}}{K e^B}=\frac{e^{-2B}}{K}.
 \label{eq:attention-share-bound}
\end{equation}
A width-$D$ interaction therefore has lower bound $e^{-2B}/D$, whereas
a width-$M$ leaf has lower bound $e^{-2B}/M$. Their ratio is $D/M$ (or
$|\mathcal C|/M$ in the Raw-Core comparison). This compares worst-case local
lower bounds; it does not bound the ratio of realized attention weights or
prediction errors because folding changes the logits and removes cross-leaf
interactions.

The resource benefit comes from replacing one large interaction map with a
sequence of small ones. For a pairwise feature mixer, one leaf costs $O(TM^2d)$. With
$G=O(D/M)$ leaves, total feature-interaction work is $O(TDMd)$ and sequential
execution bounds the live interaction map by $O(TM^2)$. The final combined
representation remains $T\times d_{\rm ICL}$, but total device memory is not constant in
$D$: the present adapters can retain $O(GTd_{\rm ICL})$ leaf representations in addition
to weights, contextual activations, and runtime workspace. We therefore treat
the measured peak-memory results as empirical systems evidence rather than as
an equality with the asymptotic interaction working set.

\subsection{Tail residual and reciprocal support-subset acceptance}

Encoding all columns solves access to evidence; it does not decide how much
of that evidence should affect a prediction. A useful Tail update should
add something beyond the Core and retain class information across support
subsets. The projection and acceptance test implement these two requirements.

Let $t=|\mathcal G_T|^{-1}\sum_{g\in\mathcal G_T}m_g$ be the averaged
Tail message and
$\operatorname{ctr}(z)=z-|S|^{-1}\sum_{i\in S}z_i$ center a representation by its
support rows. Write $b=\operatorname{ctr}(c)$ and
$u=\operatorname{ctr}(t)$. SCFF removes the global Tail direction aligned
with the Core:
\begin{equation}
  \rho=\frac{\langle u_S,b_S\rangle}{\|b_S\|_2^2},
  \qquad r=u-\rho b.
  \label{eq:tail-residual}
\end{equation}
Flatten the centered support representations $b_S$ and $u_S$. For nondegenerate $b_S$,
Equation~\eqref{eq:tail-residual} gives
\begin{equation}
 \langle r_S,b_S\rangle
 =\langle u_S,b_S\rangle
  -\frac{\langle u_S,b_S\rangle}{\|b_S\|_2^2}
   \langle b_S,b_S\rangle=0.
 \label{eq:residual-orthogonality}
\end{equation}
If the denominator is degenerate, the implementation sets the projection
coefficient to zero. Orthogonality therefore holds for the raw residual in
the nondegenerate case; it is not a claim of statistical independence. The
prototype-mapped candidate lies in the convex hull of the centered class
prototypes and hence in their span, but the nonlinear map need not preserve
the raw residual's orthogonality.

The raw residual can still contain variation unrelated to the task. The
second candidate concentrates it toward support class prototypes. For class
centroids $\mu_k$ and support class frequencies $q_k$, define
\begin{equation}
 p_k(r_i)=
 \frac{\exp(-\|r_i-\mu_k\|_2^2/s)}
      {\sum_\ell\exp(-\|r_i-\mu_\ell\|_2^2/s)},\qquad
 r_i^{\rm proto}=\sum_k p_k(r_i)
       \left(\mu_k-\sum_\ell q_\ell\mu_\ell\right),
 \label{eq:prototype-map}
\end{equation}
where $s$ is the median positive squared distance between class centroids.
For a support row, its own-class distance uses a leave-one-out centroid.
This is a support-fitted candidate, not an independently cross-fitted
representation: query rows use centroids from all support rows. Cross-half
separation and nearest-prototype checks below accept or reject the fixed candidate and
its proposed step without reading query labels.
Both the raw and prototype-mapped candidates are normalized to the support
RMS of $b$, so the update coefficient has the same scale across episodes.

For support subset $Q$, write the Fisher ratio of $b+\alpha v$ as
\begin{equation}
 J_Q(\alpha;v)=
 \frac{B_{0,Q}+2\alpha B_{1,Q}+\alpha^2B_{2,Q}}
      {W_{0,Q}+2\alpha W_{1,Q}+\alpha^2W_{2,Q}}.
 \label{eq:fisher-quadratic}
\end{equation}
The stationary points are the real roots in $[0,1]$ of a quadratic, together
with the two endpoints, so the finite candidate set gives the exact maximizer
used by the method. After splitting each class alternately into $A$ and
$B$, each half proposes a step and SCFF uses
\begin{equation}
 \alpha_v=\min(\widehat\alpha_A,\widehat\alpha_B).
 \label{eq:tail-gate}
\end{equation}
An accepted multi-leaf
update must improve Fisher separation and nearest-prototype NLL in
both directions without lowering either support accuracy. It therefore
satisfies
\begin{equation}
 J_A(\alpha_v;v)>J_A(0;v),\qquad
 J_B(\alpha_v;v)>J_B(0;v).
 \label{eq:reciprocal-fisher}
\end{equation}
If no candidate passes, $\alpha=0$ and Equation~\eqref{eq:predict-once}
returns the Bounded Core representation exactly. The single-leaf Tail fallback
uses the two-way Fisher condition alone because no cross-leaf Tail merge is
required; unlike the multi-leaf update, it does not guarantee prototype-NLL
improvement. Multiclass tasks also take the exact Core fallback.

The accepted direction is added before the original context encoder:
\begin{equation}
 h=c+\alpha v,\qquad
 p(y_q\mid X_s,y_s,X_q)=H_\theta(C_\theta(h;y_s)).
 \label{eq:appendix-predict-once}
\end{equation}
This expression also makes the exact fallback explicit: setting $\alpha=0$
returns $h=c$ while preserving one call to the context encoder and label head.

\subsection{A conditional support-to-query statement}

The remaining link is from a support-side improvement to useful evidence on
new rows. The following statement makes that link explicit and identifies
the empirical question addressed by the held-out Tail experiment.

Let $\Delta_S(v)$ be the smaller prototype-NLL gain across the two support
directions and $\Delta_Q(v)$ the corresponding population gain on unseen query
rows. The validation step establishes $\Delta_S(v)>0$ for an accepted multi-leaf update.
To connect this fact to query performance, an additional stability condition
is required:
\begin{equation}
 |\Delta_Q(v)-\Delta_S(v)|\le\varepsilon.
 \label{eq:support-query-transfer}
\end{equation}
Under this condition, any accepted update with
$\Delta_S(v)>\varepsilon$ has $\Delta_Q(v)>0$. Exchangeability motivates
estimating the query gain from support splits but does not itself imply
Equation~\eqref{eq:support-query-transfer}, nor does the proposition guarantee
improvement after an arbitrary context encoder and head. The held-out
Tail experiment in Section~\ref{app:mechanism-validation} tests precisely this
empirical transfer.

\paragraph{Finite implementation checks.}
The following tests connect the formal objects above to one executable
implementation. Their numerical tolerances verify the tested cases; they are
not substitutes for the algebraic statements or broad predictive evidence.

\begin{center}
\begin{minipage}{\linewidth}
\centering
\captionof{table}{\textbf{Executable checks for the residual/prototype update.}}
\label{tab:full-coverage-theory-controls}
\footnotesize
\setlength{\tabcolsep}{4.2pt}
\begin{tabular}{@{}lll@{}}
\toprule
Check & Setting & Observed \\
\midrule
Exact coverage & 1025 features & 1025 retained once \\
Bounded feature encoder & 9 leaves & maximum width 128 \\
Support-only routing & query representations shifted $100\times$ & validation unchanged \\
Residual geometry & accepted raw update & $|\langle r_S,b_S\rangle|=5.1e-07$ \\
Prototype geometry & rank-1 class span & projection error 6.0e-08 \\
Split-half validation & support halves $A,B$ & Fisher $0.080\to2.684$ \\
Prototype prediction & both directions & minimum NLL gain 0.439 \\
Nuisance width & 2--16 symmetric Tail leaves & maximum Core change 0.0e+00 \\
Exact fallback & rejected binary / multiclass Tail & weights 0/0, bit-exact \\
\bottomrule
\end{tabular}
\end{minipage}
\end{center}

\section{Implementation and Architecture Interfaces}
\label{app:implementation}

The framework depends on where a model permits evidence to be combined.
This section follows a table through support compilation and the architecture
adapter, showing how the bounded-computation idea is realized while preserving
the downstream prediction interface.

\subsection{Support compiler}

The method operates on a median-filled copy of the support features.
It computes one-way ANOVA statistics, uses the Benjamini--Hochberg discovery
count at nominal FDR $0.05$ to determine Core size, and assigns the
highest-$F_j$ columns to the Core. At least one column initializes a nonempty
Core. ANOVA/BH therefore routes columns; it does not delete the Tail. The
model still receives its native preprocessing and missing-value
handling. Every leaf uses the same feature encoder and receives all support
and query rows. Core and Tail leaf representations are averaged with equal leaf weight
before support-only validation is applied.

The compatible insertion point differs across architectures. Table~\ref{tab:evidence-tiers}
records the comparison and baseline for each reported model; these
distinctions prevent a capacity extension from being interpreted as a
matched-width efficacy result.

\begin{table}[h]
\centering
\caption{\textbf{Model-specific evaluation settings.} Matched-width
comparisons use the same features, comparisons within native limits respect
each model's output constraints, and capacity extensions use wider inputs than
the original model.}
\label{tab:evidence-tiers}
\small
\setlength{\tabcolsep}{4pt}
\begin{tabularx}{\linewidth}{@{}lYll@{}}
\toprule
Model & Comparison & Coverage & Interpretation \\
\midrule
TabICLv2 & matched full width & 90 folds / 18 datasets & direct comparison \\
TabPFN-3 & matched full width & 90 folds / 18 datasets & direct comparison \\
TabICLv1 & matched full width & 90 folds / 18 datasets & direct comparison \\
RefineICL-24L & matched full width & 90 folds / 18 datasets & direct comparison \\
TabPFNv1 & support-ranked 128 vs. native 100-feature view & 85 folds / 17 datasets & capacity extension \\
TabPFNv2 & latent representation interface & standard-path test & representation boundary \\
LimiX-2M & token-aligned representation combination & 85 folds / 17 datasets & LimiX adapter \\
\bottomrule
\end{tabularx}
\end{table}

\begin{table}[h]
\centering
\caption{Execution checks for every reported result.}
\label{tab:nonensemble}
\small
\begin{tabular}{@{}lr@{}}
\toprule
Quantity & Required value \\
\midrule
Model checkpoints & 1 \\
Feature partitions from support data & 1 \\
Context-encoder calls per query & 1 \\
Label-head calls per query & 1 \\
Averaged final predictions & 0 \\
Query labels used to build feature groups & false \\
Feature budget reduced after OOM & false \\
\bottomrule
\end{tabular}
\end{table}

\subsection{Row-representation and token-preserving adapters}

TabPFN-3, TabICLv1/v2, TabPFNv1, and RefineICL-24L expose compatible row representations
before a shared context encoder. Their adapters combine aligned
$T\times d_{\rm ICL}$ messages. TabPFNv1's standard inference uses a fixed
100-feature view, so its 17-dataset comparison asks whether support compilation
extends usable capacity rather than matching full input width.

\paragraph{LimiX token adapter.}
\label{app:limix-token-adapter}
LimiX forms one token from each native group of $g$ scalar columns before its
alternating feature--row transformer. Averaging at the row-representation interface used by
the other models would destroy this trained token axis. The LimiX variant
therefore encodes leaves of at most $M$ columns with the released tokenizer,
padding the last incomplete native group of each leaf. Tokens follow compiled
Core-first leaf order, with original column order within each leaf.
Let $w_{\max}=\max_j|P_j|$ be the actual maximum leaf width.
For $T_f$ resulting feature tokens, the output budget is
\begin{equation}
B=\min\!\left(T_f,\left\lceil w_{\max}/g\right\rceil\right),\quad
B_{\mathrm C}=\min\!\left(T_f-1,
\max\!\left(1,\operatorname{round}(0.75B)\right)\right),\quad
B_{\mathrm T}=B-B_{\mathrm C}.
\end{equation}
When $T_f>B$, the first $B_{\mathrm C}$ Core tokens remain unchanged and the
remaining tokens are split evenly across $B_{\mathrm T}$ Tail slots, averaged,
and support-RMS rescaled. For the reported LimiX-2M experiments, $g=2$ and
$M=512$; 192 Core and 64 Tail slots result only when $B=256$.
Smaller leaves use correspondingly smaller budgets. When $T_f\le B$, all
encoded tokens are retained. A one-leaf plan bypasses compilation and calls
native inference exactly. This token-preserving combination is
training-free and single-prediction, but it does not use the split-half
row-representation Tail update analyzed above.

\subsection{Verifying one complete prediction}

Repeated feature-encoder calls are essential to folding. The following checks
ensure that they collectively supply one prediction and that the measured
savings do not arise from lost columns or altered evaluation data.

For every reported evaluation, sorted leaf indices must equal $0,\ldots,D-1$ without
duplication; each leaf must respect its width bound; and the plan and Tail validation
must read support features and labels only. A rejected Tail update must reproduce
the exact Bounded Core representation. Each query uses one checkpoint member, one
feature partition, one context-encoder call, one label-head call, and one
complete class distribution. Leaf-local prototype scores select a candidate Tail
update and are never combined as final predictions. Query labels are
read only after the predictions have been produced.

\section{Data and Statistical Protocol}
\label{app:protocol}

To test width adaptation, the evaluation population must be defined by width
before observing model gains. To test the method itself, native and folded
inference must then see the same task. The registry and pairing rules below
establish these two conditions; the locked-fold analyses assess whether the
selected design continues to help on new query rows.

\subsection{Benchmark selection}

The OpenML Wide-Table panel is the exhaustive set of dataset versions in the
fixed AMLB-29, released TabZilla, and TabArena v0.1 snapshots satisfying the
eligibility rule, which uses input metadata only. Exact OpenML data IDs identify versions and semantic
duplicates are excluded before outcomes are observed. The dataset population
itself is outcome-blind; fold 0 of this fixed population is nevertheless used
for the finite method-selection procedure described below.
This is a controlled mechanism benchmark assembled from public registries, not
an official leaderboard aggregate.

\begin{table*}[t]
\centering
\caption{\textbf{Datasets in the OpenML Wide-Table Holdout.}
The population is the exhaustive set of classification datasets in the public
fixed registry snapshot from September 4, 2026 covering AMLB-29, the released
TabZilla task collection, and TabArena v0.1 that satisfy the preregistered
metadata rule: at least 128 raw features, 2--50
classes, no missing target, and no exact data ID or semantic reissue from the
earlier exploratory development set. Fold 0 of this fixed panel was then used
for the declared finite method selection; folds 1--4 were reserved for
locked-method evaluation. OpenML data IDs identify the immutable dataset versions;
no model prediction or query label enters dataset selection. ``TabZilla list''
denotes the paper's released full task collection, while 379 and 457 denote
the corresponding OpenML benchmark suites.}
\label{tab:public-wide-datasets}
\scriptsize
\setlength{\tabcolsep}{4.0pt}
\begin{tabular}{@{}rlrrrr@{}}
\toprule
OpenML data ID & Dataset version & Benchmark source & Rows & Features & Classes \\
\midrule
5     & arrhythmia                         & TabZilla list                 & 452    & 279   & 13 \\
312   & scene                              & TabZilla list                 & 2,407  & 299   & 2 \\
1036  & sylva\_agnostic                    & TabZilla list                 & 14,395 & 216   & 2 \\
1038  & gina\_agnostic                     & TabZilla list                 & 3,468  & 970   & 2 \\
1116  & musk                               & TabZilla list                 & 6,598  & 169   & 2 \\
1476  & gas-drift                          & TabZilla list                 & 13,910 & 128   & 6 \\
1477  & gas-drift-different-concentrations & TabZilla list                 & 13,910 & 129   & 6 \\
41142 & christine                          & TabZilla list                 & 5,418  & 1,636 & 2 \\
41143 & jasmine                            & AMLB-29; TabZilla 379/list    & 2,984  & 144   & 2 \\
41144 & madeline                           & AMLB-29                       & 3,140  & 259   & 2 \\
41145 & philippine                         & AMLB-29; TabZilla list        & 5,832  & 308   & 2 \\
41159 & guillermo                          & TabZilla 379/list             & 20,000 & 4,296 & 2 \\
41161 & riccardo                           & TabZilla list                 & 20,000 & 4,296 & 2 \\
41163 & dilbert                            & TabZilla list                 & 10,000 & 2,000 & 5 \\
41164 & fabert                             & TabZilla list                 & 8,237  & 800   & 7 \\
41165 & robert                             & TabZilla list                 & 10,000 & 7,200 & 10 \\
46933 & hiva\_agnostic                     & TabArena v0.1 / suite 457     & 3,845  & 1,617 & 3 \\
46939 & kddcup09\_appetency                & TabArena v0.1 / suite 457     & 50,000 & 212   & 2 \\
\bottomrule
\end{tabular}
\end{table*}

\FloatBarrier

\begin{center}\begin{minipage}{0.98\linewidth}
\centering
\captionof{table}{\textbf{Concrete locked-fold reproduction contract.} Values
are read from the released driver and per-fold receipts; hash prefixes identify
the exact frozen artifacts.}
\label{tab:reproduction-contract}
\footnotesize
\setlength{\tabcolsep}{4pt}
\begin{tabularx}{\linewidth}{@{}lY@{}}
\toprule
Item & Frozen value \\
\midrule
Population & 18 listed OpenML versions (17 for TabPFNv1); selected from registry metadata before model outcomes \\
Fold construction & deterministic stratified 5-fold; split seed 20260904; fold 0 development, folds 1--4 locked confirmation \\
Overlap & query folds are pairwise disjoint; each support set is the complement and therefore overlaps other support sets \\
Caps and sampling & support $\le2048$, query $\le1024$; stratified cap seed derived per dataset; query subsampling seed 20260905 \\
Method & ANOVA/BH FDR 0.05; $M=128$; balanced round-robin Core/Tail leaves; final residual/prototype gate \\
Execution & one member, one preprocessing view, one context-stack call and one label-head call; no prediction averaging \\
OOM policy & halve the paired query chunk on both paths only; never reduce support rows or features; exclude incomplete pairs \\
Hardware and memory & PPU-ZW810 (72 SM, CUDA-compatible); both arms paired on the same device; peak allocated bytes from PyTorch \\
Precision & released model-native inference precision; native and SCFF share the same checkpoint, runtime and preprocessing \\
TabPFN-3 & checkpoint \texttt{d0d865d54dfb}; source \texttt{b03a4f9547bd}; temperature 0.9 \\
TabPFN-3 view & one full-width \texttt{quantile\_uni}/numeric transform; no feature or class shift, fingerprint, polynomial expansion, or within-registry feature subsampling \\
TabICLv1 & checkpoint \texttt{04c5c1d261c1}; source \texttt{59a957cd644b}; temperature 0.9 \\
TabICLv2 & checkpoint \texttt{bdc7dbd5e4ff}; source \texttt{59a957cd644b}; temperature 0.9 \\
RefineICL-24L & step-15000 checkpoint \texttt{071de368d16d}; source \texttt{2d0bb39ce7fa}; temperature 0.9 \\
TabPFNv1 extension & checkpoint \texttt{3c9aadaeddbf}; source \texttt{a9298ae0fe2a}; temperature 0.8 \\
Protocol hashes & manifest \texttt{e9339459ecb8}; evaluation \texttt{fdbf4d9345d2}; statistics \texttt{56d8255b4a95} \\
\bottomrule
\end{tabularx}
\end{minipage}\end{center}

The five query folds form a partition of each dataset under deterministic
stratified splitting. Consequently, fold 0 query rows never occur as query
rows in folds 1--4, although they enter those folds' support sets, as standard
cross-validation requires. The confirmation therefore tests new query rows
after freezing the method, not transfer to unseen datasets. Dataset-bootstrap
intervals estimate variation across the fixed dataset population; they do not
turn this protocol into leave-dataset-out method development.

\subsection{Pairing, aggregation, and resources}

A paired evaluation fixes dataset version, fold, support/query indices, preprocessing,
checkpoint, query chunking, and runtime. Both methods must complete without
fallback or replacement. For metric $m$, we first average fold-level paired
effects within dataset $d$ and then macro-average dataset effects,
\begin{equation}
 \Delta m=\frac{1}{|\mathcal D|}\sum_{d\in\mathcal D}
 \left(\frac{1}{|\mathcal F_d|}\sum_{f\in\mathcal F_d}
 [m_{d,f}^{\rm SCFF}-m_{d,f}^{\rm native}]\right).
 \label{eq:paired-aggregation}
\end{equation}
The headline table contains fold 0 as well as locked folds 1--4; its point
estimates therefore describe the fixed five-fold panel, not an untouched test
of method selection. We therefore confirm the complete method separately on
folds 1--4 and exclude fold 0 from that analysis.
Unless an interval is printed, the table reports a point estimate rather than
statistical significance.

We additionally resample datasets with replacement after the within-dataset
fold average, preserving the dataset as the unit of generalization. All four
matched-width comparisons have favorable locked-fold 95\% intervals for both
metrics. The token-aligned comparison is not reconstructed under this fold
contract; the capacity extension remains a positive point estimate whose
interval crosses zero.

\begin{center}\begin{minipage}{0.94\linewidth}
\centering
\captionof{table}{\textbf{Locked-fold confirmation of the complete SCFF method.} The method selected on fold 0 is evaluated only on folds 1--4. Effects are averaged within dataset before 100,000 paired dataset-bootstrap resamples. $\ddagger$ denotes the separately declared capacity extension.}
\label{tab:heldout-full-method}
\footnotesize\setlength{\tabcolsep}{4pt}
\begin{tabular}{@{}lrrr@{}}
\toprule
Backbone & Data & $\Delta$Accuracy pp [95\% CI] & $\Delta$NLL [95\% CI] \\
\midrule
TabPFN-3 & 18 & +1.956 [+0.212, +4.808] & -0.05103 [-0.11702, -0.00816] \\
TabICLv1 & 18 & +5.348 [+1.663, +9.846] & -0.08828 [-0.15570, -0.03260] \\
TabICLv2 & 18 & +1.615 [+0.546, +2.999] & -0.03603 [-0.06176, -0.01520] \\
RefineICL-24L & 18 & +3.846 [+0.323, +9.136] & -0.08394 [-0.17533, -0.01588] \\
TabPFNv1$^\ddagger$ & 17 & +3.063 [-0.962, +7.368] & -0.04479 [-0.11306, +0.02135] \\
\bottomrule
\end{tabular}
\end{minipage}\end{center}

Positive averages also invite a distributional question: does the conclusion
depend on one exceptional dataset? The leave-one-dataset-out analysis below
removes each dataset in turn from the locked-fold comparison. Favorable
effects remain on all four matched-width backbones, complementing the
dataset-bootstrap intervals with a direct check of individual influence.

\begin{center}\begin{minipage}{0.96\linewidth}
\centering
\captionof{table}{\textbf{Locked-fold gains survive every single-dataset deletion.} For each matched-width backbone, we remove each dataset in turn and report the least favorable remaining macro effect for each metric. The named dataset is the deletion that produces that worst case.}
\label{tab:heldout-influence}
\footnotesize\setlength{\tabcolsep}{4pt}
\begin{tabular}{@{}lrrll@{}}
\toprule
Backbone & Full $\Delta$Acc. & Min LODO $\Delta$Acc. & Full $\Delta$NLL & Max LODO $\Delta$NLL \\
\midrule
TabPFN-3 & +1.956 & +0.689 (fabert) & -0.05103 & -0.02276 (fabert) \\
TabICLv1 & +5.348 & +3.831 (madeline) & -0.08828 & -0.06831 (fabert) \\
TabICLv2 & +1.615 & +1.133 (guillermo) & -0.03603 & -0.02690 (robert) \\
RefineICL-24L & +3.846 & +1.595 (riccardo) & -0.08394 & -0.04775 (riccardo) \\
\bottomrule
\end{tabular}
\par\smallskip\footnotesize LODO: leave one dataset out. Positive Accuracy and negative NLL favor SCFF.
\end{minipage}\end{center}

\begin{center}\begin{minipage}{0.92\linewidth}
\centering
\captionof{table}{\textbf{Paired dataset-bootstrap intervals for the headline panel.} Effects are computed after averaging folds within each dataset; brackets are 95\% percentile intervals over 100,000 dataset resamples. $\dagger$ is token-aligned and $\ddagger$ is a capacity extension; the remaining four rows are matched-width comparisons.}
\label{tab:headline-bootstrap}
\footnotesize
\setlength{\tabcolsep}{4pt}
\begin{tabular}{@{}lrrr@{}}
\toprule
Backbone & Datasets & $\Delta$Accuracy pp [95\% CI] & $\Delta$NLL [95\% CI] \\
\midrule
TabPFN-3 & 18 & +1.874 [+0.036, +4.806] & -0.05065 [-0.11723, -0.00654] \\
TabICLv1 & 18 & +5.419 [+1.727, +9.977] & -0.09099 [-0.15981, -0.03462] \\
TabICLv2 & 18 & +1.564 [+0.528, +2.918] & -0.03557 [-0.06120, -0.01496] \\
RefineICL-24L & 18 & +3.730 [+0.380, +8.764] & -0.08223 [-0.16831, -0.01732] \\
LimiX-2M$^\dagger$ & 17 & +0.700 [-0.769, +2.338] & -0.01178 [-0.03719, +0.01125] \\
TabPFNv1$^\ddagger$ & 17 & +3.071 [-0.957, +7.297] & -0.04341 [-0.11047, +0.02201] \\
\bottomrule
\end{tabular}
\end{minipage}\end{center}

End-to-end time sums prediction and support compilation over complete units.
Peak allocated memory is measured by the PyTorch CUDA allocator. In the
text's maximum-peak comparison, each backbone's ratio divides its largest
observed native peak by its largest observed SCFF peak; those maxima need
not occur on the same fold. It is not the maximum of paired-fold ratios.
Feature partitions may be reused for repeated query
batches, but the reported end-to-end number charges compilation once.

\begin{center}\begin{minipage}{0.9\linewidth}
\centering
\captionof{table}{\textbf{Paired resource ratios across dataset-fold runs.} Each ratio compares native and SCFF on the same query fold. Memory is native/SCFF peak allocated GPU memory; speed is native/SCFF end-to-end time including support compilation. Medians and 10th--90th percentiles describe folds, not ratios of cross-task extrema; only directly measured or hash-verified native references are included. RefineICL has 90 predictive folds but only 72 resource pairs: all 18 fold-2 native reference file hashes differ from the retained baseline artifact and are excluded rather than imputed.}
\label{tab:paired-resource-distribution}
\footnotesize\setlength{\tabcolsep}{4pt}
\begin{tabular}{@{}lrrr@{}}
\toprule
Backbone & Pairs & Memory, median [P10, P90] & Speed, median [P10, P90] \\
\midrule
TabPFN-3 & 90 & 2.15$\times$ [1.36, 21.99] & 1.00$\times$ [0.63, 2.04] \\
TabICLv1 & 90 & 2.36$\times$ [1.56, 25.39] & 0.62$\times$ [0.40, 0.96] \\
TabICLv2 & 90 & 2.32$\times$ [1.54, 27.29] & 0.56$\times$ [0.33, 0.93] \\
RefineICL-24L & 72 & 2.09$\times$ [1.23, 23.90] & 4.84$\times$ [3.13, 6.74] \\
LimiX-2M & 85 & 2.46$\times$ [1.02, 4.84] & 1.72$\times$ [0.69, 5.61] \\
TabPFNv1 & 85 & 0.89$\times$ [0.30, 0.97] & 0.18$\times$ [0.06, 1.17] \\
\bottomrule
\end{tabular}
\end{minipage}\end{center}

For every result, we record the checkpoint SHA-256, source revision, dataset,
fold, support/query hashes, preprocessing, method hash, adapter revision, and
runtime. Changing the router, FDR threshold, leaf width,
residual map, or validation rule defines a new protocol.

\section{Mechanism Validation}
\label{app:mechanism-validation}

The main table shows what SCFF achieves as a complete method. The experiments
here ask why. We first hold the feature set fixed, then examine where folding
changes the representation, and finally test whether weaker evidence can
contribute beyond the resulting Core. This progression assigns distinct roles
to selection, bounded computation, and Tail recovery.

\subsection{Why separate Core folding from Tail recovery?}

The headline comparison changes both the representation of selected evidence
and the treatment of lower-ranked columns. Two paired controls identify these
effects. Raw Core and Bounded Core receive the same ranked feature set; only
the latter uses bounded leaves. Tail Off and Tail On share the exact Bounded
Core representation; only Tail On executes the remaining features and adds a support-checked Tail residual.
Thus the first contrast tests bounded local competition, while the second asks
whether evidence outside the Core passes a reciprocal support-subset acceptance test.
The component ablation in main Table~\ref{tab:full-coverage-component-ablation}
uses the same paired units. Its coupled controls are not additive effects.

\begin{center}\begin{minipage}{0.92\linewidth}
\centering
\captionof{table}{\textbf{Uncertainty for the full-panel Core decomposition.} Selection-only minus native and Bounded Core minus selection-only over all five folds; brackets are 95\% paired dataset-bootstrap intervals after averaging folds within each dataset.}
\label{tab:core-confirmation-intervals}
\footnotesize
\setlength{\tabcolsep}{5.5pt}
\begin{tabular}{@{}llrr@{}}
\toprule
Backbone & Contrast & $\Delta$Acc. pp $\uparrow$ & $\Delta$NLL $\downarrow$ \\ 
\midrule
TabPFN-3 & Selection $-$ native & \shortstack{+1.563\\[-1pt]{\scriptsize [-0.197, +4.481]}} & \shortstack{-0.03414\\[-1pt]{\scriptsize [-0.09835, +0.00407]}} \\
 & Bounded $-$ selection & \shortstack{+0.233\\[-1pt]{\scriptsize [-0.094, +0.730]}} & \shortstack{-0.00639\\[-1pt]{\scriptsize [-0.01629, +0.00042]}} \\
TabICLv1 & Selection $-$ native & \shortstack{+5.175\\[-1pt]{\scriptsize [+1.465, +9.716]}} & \shortstack{-0.07959\\[-1pt]{\scriptsize [-0.14791, -0.02282]}} \\
 & Bounded $-$ selection & \shortstack{+0.162\\[-1pt]{\scriptsize [-0.284, +0.683]}} & \shortstack{-0.00911\\[-1pt]{\scriptsize [-0.02096, +0.00105]}} \\
TabICLv2 & Selection $-$ native & \shortstack{+0.690\\[-1pt]{\scriptsize [-0.155, +1.884]}} & \shortstack{-0.01576\\[-1pt]{\scriptsize [-0.03360, -0.00119]}} \\
 & Bounded $-$ selection & \shortstack{+0.739\\[-1pt]{\scriptsize [+0.117, +1.656]}} & \shortstack{-0.01683\\[-1pt]{\scriptsize [-0.04114, -0.00048]}} \\
\bottomrule
\end{tabular}
\vspace{2pt}

\parbox{0.98\linewidth}{\footnotesize Selection supplies most of the quality gain. Holding that selected Core fixed, bounded folding improves both point estimates on all three backbones; the ratio of separate maximum Core peaks is 30.1--34.0$\times$, while both folding intervals exclude zero only on TabICLv2. Thus the robust folding claim is capacity preservation with a consistent, smaller quality increment, not that folding explains the full native-to-SCFF gain.}
\end{minipage}\end{center}

Fold 0 was used for development. Recomputing the fixed-Core comparison on
query folds 1--4 alone gives the secondary paired analysis in
Table~\ref{tab:heldout-core-quality}. This contrast was examined after the
five-fold analysis and uses new query rows from the same datasets, not new
datasets. It separates the development fold without changing selected
columns, checkpoints, or the two paired prediction routes.

\begin{center}\begin{minipage}{0.92\linewidth}
\centering
\captionof{table}{\textbf{Locked-query-fold same-Core contrast.} Bounded Core minus Raw Core on folds 1--4; both use the same selected columns, checkpoint, support and query indices. Positive accuracy difference and NLL reduction favor bounded folding. Brackets are 95\% dataset-cluster bootstrap intervals over 18 datasets, averaging four folds within dataset. This secondary contrast was examined after the all-five-fold analysis; folds are new query rows from the same datasets, not unseen datasets.}
\label{tab:heldout-core-quality}
\footnotesize\setlength{\tabcolsep}{5pt}
\begin{tabular}{@{}lrr@{}}
\toprule
Backbone & $\Delta$Acc. pp $\uparrow$ & NLL reduction $\uparrow$ \\
\midrule
TabPFN-3 & \shortstack{+0.308\\[-1pt]{\scriptsize [+0.005, +0.808]}} & \shortstack{+0.00744\\[-1pt]{\scriptsize [+0.00047, +0.01802]}} \\
TabICLv1 & \shortstack{+0.171\\[-1pt]{\scriptsize [-0.277, +0.660]}} & \shortstack{+0.00851\\[-1pt]{\scriptsize [-0.00283, +0.02134]}} \\
TabICLv2 & \shortstack{+0.768\\[-1pt]{\scriptsize [+0.156, +1.702]}} & \shortstack{+0.01787\\[-1pt]{\scriptsize [+0.00084, +0.04291]}} \\
\midrule
Equal-model mean & \shortstack{+0.416\\[-1pt]{\scriptsize [+0.099, +0.842]}} & \shortstack{+0.01127\\[-1pt]{\scriptsize [+0.00111, +0.02591]}} \\
\bottomrule
\end{tabular}
\end{minipage}\end{center}

\subsection{When does bounded Core processing change predictions?}

The mechanism predicts where a difference should emerge: below one leaf,
the two routes should nearly coincide; above two
leaf widths, local feature interactions differ substantially. We partition
folds by the selected Core width $K$ under the fixed $M=128$ cap, average
paired differences within each dataset, and then macro-average datasets in
each bin. Table~\ref{tab:core-width-quality} reports every bin for all three
latent-state backbones. At $K>256$, both quality point estimates favor bounded
Core on each backbone; $K\leq128$ is approximately an identity control.
Only five datasets contribute to the widest bin per backbone, however, and
this stratification was examined after evaluation. It describes where the
observed effect lies, not an independently confirmed width threshold or a
guarantee of predictive improvement.

\begin{center}\begin{minipage}{0.92\linewidth}
\centering
\captionof{table}{\textbf{Exploratory same-Core quality by selected width.} Raw and bounded Core share columns, checkpoint, and query fold. Within each width bin, folds are averaged within dataset before dataset-macro aggregation. Positive values favor bounded Core. Bins are defined by the $M=128$ leaf cap; datasets can contribute different folds to adjacent bins. These descriptive contrasts were examined after the main evaluation, not preregistered as confirmatory tests.}
\label{tab:core-width-quality}
\footnotesize\setlength{\tabcolsep}{5pt}
\begin{tabular}{@{}llrrrr@{}}
\toprule
Backbone & Core $K$ & Datasets & Folds & $\Delta$Acc. pp $\uparrow$ & NLL reduction $\uparrow$ \\
\midrule
TabPFN-3 & $K\leq128$ & 9 & 45 & +0.000 & +0.00000 \\
TabPFN-3 & $129$--$256$ & 5 & 21 & -0.316 & -0.00268 \\
TabPFN-3 & $K>256$ & 5 & 24 & +1.084 & +0.02545 \\
\addlinespace
TabICLv1 & $K\leq128$ & 10 & 44 & +0.000 & +0.00000 \\
TabICLv1 & $129$--$256$ & 6 & 22 & +0.137 & +0.00178 \\
TabICLv1 & $K>256$ & 5 & 24 & +0.714 & +0.03472 \\
\addlinespace
TabICLv2 & $K\leq128$ & 10 & 44 & +0.000 & +0.00000 \\
TabICLv2 & $129$--$256$ & 6 & 22 & +1.215 & +0.03033 \\
TabICLv2 & $K>256$ & 5 & 24 & +1.917 & +0.04975 \\
\bottomrule
\end{tabular}
\end{minipage}\end{center}

\subsection{How was the Tail update selected and confirmed?}

The Tail presents a harder design problem than the Core: more encoded
information may be redundant or may obscure a useful representation. We
therefore compare ways of proposing and accepting that information before
testing the chosen rule on new query rows. Fold 0 compared four specified Tail updates across all five
compatible backbones. Selection first maximized the number of backbones with
simultaneous dataset-macro accuracy and NLL improvement, then compared mean NLL
and accuracy. This is development evidence over a finite candidate set, not a
claim that all possible Tail updates were searched.

\begin{center}\begin{minipage}{\linewidth}
\centering
\captionof{table}{\textbf{Selecting the Tail update across models.} Each variant is evaluated on fold 0 over all five compatible models. Selection first maximizes the number of models with simultaneous accuracy and NLL improvement, then compares mean NLL and accuracy.}
\label{tab:tail-operator-development}
\footnotesize
\setlength{\tabcolsep}{5.0pt}
\begin{tabular}{@{}lrrr@{}}
\toprule
Tail update & Passing models & Mean $\Delta$Acc. pp & Mean $\Delta$NLL \\
\midrule
Raw residual update & 2/5 & $+0.090$ & $-0.00130$ \\ 
Cross-fitted residual update & 2/5 & $+0.079$ & $-0.00172$ \\ 
Centered cross-fitted update & 2/5 & $+0.080$ & $-0.00163$ \\ 
\textbf{Full residual/prototype validation} & \textbf{3/5} & $\mathbf{+0.025}$ & $\mathbf{-0.00304}$ \\ 
\bottomrule
\end{tabular}
\vspace{2pt}

\parbox{0.98\linewidth}{\footnotesize Each model effect is a dataset macro over the 18 predeclared wide datasets (17 for TabPFNv1). The selected rule is the only complete variant that improves both metrics on at least three of five models. Its implementation was fixed before folds 1--4 were evaluated.}
\end{minipage}\end{center}

The selected implementation was fixed before folds 1--4 were evaluated.
The confirmation comprises 72 paired units for each 18-dataset backbone and 68
for TabPFNv1's 17-dataset panel within its native limits. Three of five backbones improve
both metrics, satisfying the predeclared three-of-five rule. Held-out transfer is not universal:
TabPFNv1 loses $0.007$ accuracy points despite a negligible
NLL gain, and RefineICL-24L degrades both metrics. The five-backbone comparison
therefore supports a consistent three-backbone latent-state result plus a
documented interface boundary, not universal Tail improvement or
backbone-level statistical significance.

Paired dataset-bootstrap intervals quantify effect strength rather than replace
the predeclared cross-backbone criterion. TabICLv2 has favorable intervals for
both Tail effects: Accuracy $+0.145$ pp $[+0.009,+0.334]$ and NLL $-0.00294$
$[-0.00673,-0.00012]$; the other backbones' smaller Tail intervals cross zero
for at least one metric. Thus simultaneous Accuracy and NLL improvement on all
three representative latent-state backbones is the main Tail result, while
TabICLv2 provides the strongest dataset-level interval evidence.

\begin{center}\begin{minipage}{\linewidth}
\centering
\captionof{table}{\textbf{Complete Tail-extraction confirmation.} Tail On minus Tail Off over held-out folds 1--4 for every compatible backbone.}
\label{tab:tail-confirmation-complete}
\footnotesize
\setlength{\tabcolsep}{7.0pt}
\begin{tabular}{@{}lrr@{}}
\toprule
Backbone & $\Delta$Acc. pp $\uparrow$ & $\Delta$NLL $\downarrow$ \\
\midrule
TabPFN-3 & +0.107 & -0.00985 \\
TabICLv1 & +0.080 & -0.00234 \\
TabICLv2 & +0.145 & -0.00294 \\
TabPFNv1 & -0.007 & -0.00007 \\
RefineICL-24L & -0.299 & +0.00559 \\
\bottomrule
\end{tabular}
\vspace{2pt}

\parbox{0.98\linewidth}{\footnotesize The selected Tail update improves both primary metrics on three of five compatible models, satisfying the predeclared selection rule.}
\end{minipage}\end{center}

\subsection{What do simpler and resource-matched baselines explain?}
\label{app:simple-baselines}

Memory efficiency becomes useful when it changes what evidence can be
retained. These controls separate that payoff from ordinary feature
selection. The same-$K$ comparisons ask whether
bounded leaves preserve a fixed selected set while changing its resource
profile. The measured-memory comparison then gives selection-only the widest
feasible single leaf under the folded route's observed peak budget. It uses
the fixed 18-dataset registry and held-out folds 1--4; widths are selected by
memory alone, before prediction metrics are aggregated.

\begin{center}\begin{minipage}{\linewidth}
\centering
\captionof{table}{\textbf{Folding buys capacity with a measured latency cost.} The retained columns, checkpoint, fit, and complete-label prediction budget are identical between a single unbounded leaf and leaf-64 \scff{}. Times are medians of three synchronized candidate forwards after one warmup; rows report medians over paired units. Speed is single-leaf time divided by multi-leaf time, and memory is the multi/single ratio of incremental peak CUDA allocation.}
\label{tab:folding-resource}
\small
\setlength{\tabcolsep}{4.5pt}
\begin{tabular}{@{}lrrrrrr@{}}
\toprule
Dataset & Units & Retained $K$ & Single ms $\downarrow$ & Folded ms $\downarrow$ & Speed $\uparrow$ & Memory $\downarrow$ \\
\midrule
Internet-Advertisements & 10 & 512 & 404.2 & 667.3 & 0.597$\times$ & 0.134$\times$ \\
Bioresponse & 9 & 768--832 & 776.9 & 950.1 & 0.781$\times$ & 0.093$\times$ \\
cnae-9 & 10 & 256 & 127.0 & 197.3 & 0.654$\times$ & 0.259$\times$ \\
gina\_agnostic & 1 & 448 & 302.4 & 419.0 & 0.722$\times$ & 0.149$\times$ \\
\midrule
All units & 30 & 256--832 & 401.6 & 590.3 & 0.709$\times$ & 0.134$\times$ \\
\bottomrule
\end{tabular}
\vspace{2pt}
\parbox{0.98\linewidth}{\footnotesize All 30 units complete with zero failure. Prediction and retained-feature hashes are checked before aggregation. Folding lowers incremental peak allocation on 30/30 units, to 13.4\% of the single-leaf route at the overall median. It is slower on 28/30 units: median same-feature speed is 0.709$\times$ (about 1.41$\times$ longer). At $K>512$, the median memory ratio is 0.093$\times$.}
\end{minipage}\end{center}

\begin{center}\begin{minipage}{\linewidth}
\centering
\captionof{table}{\textbf{Selected evidence under the same measured peak-memory budget.} TabICLv2, one member, $M=128$, four held-out folds on each dataset in the fixed 18-dataset registry. $K$ is the support-selected Core width; $k$ is the widest single-leaf ANOVA prefix fitting the paired folded peak. Positive accuracy and NLL reduction favor folding.}
\label{tab:equal-memory-confirmation}
\footnotesize
\setlength{\tabcolsep}{4pt}
\begin{tabular}{@{}lrrrr@{}}
\toprule
Dataset & Median $K$ & Median $k$ & $\Delta$Acc. pp & NLL reduction \\
\midrule
Arrhythmia & 186 & 64 & +0.00 & -0.0223 \\
Scene & 242 & 96 & +0.00 & +0.0003 \\
Sylva & 54 & 54 & 0.00 & 0.0000 \\
Gina & 391 & 112 & +2.34 & +0.0608 \\
Musk & 129 & 80 & +3.71 & +0.1187 \\
Gas Drift & 127 & 96 & -0.02 & +0.0007 \\
Gas Drift (concentrations) & 129 & 64 & -0.10 & -0.0007 \\
Christine & 684 & 96 & +0.59 & +0.0038 \\
Jasmine & 64 & 64 & -0.13 & -0.0002 \\
Madeline$^\dagger$ & 13 & 13 & 0.00 & 0.0000 \\
Philippine & 117 & 117 & +0.54 & +0.0057 \\
Guillermo & 55 & 55 & 0.00 & 0.0000 \\
Riccardo & 357 & 96 & +1.54 & +0.0358 \\
Dilbert & 1,691 & 128 & +3.47 & +0.1021 \\
Fabert & 229 & 96 & +12.43 & +0.2710 \\
Robert & 6,753 & 128 & +16.63 & +0.4328 \\
Hiva & 1 & 1 & 0.00 & 0.0000 \\
KDDCup09 Appetency$^\dagger$ & 5 & 5 & 0.00 & 0.0000 \\
\bottomrule
\end{tabular}
\vspace{2pt}

\parbox{0.98\linewidth}{\footnotesize Dataset-macro differences average the four paired folds. All 18 datasets and 428 width points completed numerically. In 66/72 folds, $k$ satisfies the strict measured-memory inequality. $^\dagger$The other six folds have $K=k\leq14$ and identical predictions on the two one-leaf routes; single-run peak readings differ by 0.19--1.39\%. The strict complete-case estimate retains the 16 datasets with all four folds within budget (64 folds), excluding both marked datasets. The $K>M$ stratum in the main text contains 10 datasets and all 40 of its folds satisfy the measured budget.}
\end{minipage}\end{center}

The same frozen memory-only prefix grid confirms the capacity comparison on
TabPFN-3 under its strict one-member inference contract. All 430 prefix
configurations and 72 bounded-Core evaluations completed on the same host;
each fold had a feasible single-leaf prefix at or below its bounded peak.
The bounded-Core predictions exactly reproduce the prior held-out receipt.
Table~\ref{tab:equal-memory-tabpfn3} reports both the complete registry and
the width stratum fixed before inspecting outcomes.

\begin{center}\begin{minipage}{\linewidth}
\centering
\captionof{table}{\textbf{TabPFN-3 Core folding at the bounded route's measured peak-memory budget.} The single-leaf control uses the widest support-ranked Core prefix in the fixed grid whose measured peak does not exceed the paired bounded-Core peak. Positive accuracy and NLL reduction favor folding. Intervals resample datasets after averaging the four held-out folds.}
\label{tab:equal-memory-tabpfn3}
\footnotesize
\setlength{\tabcolsep}{4pt}
\begin{tabular}{@{}lrrll@{}}
\toprule
Scope & Datasets & Folds & $\Delta$Acc. pp [95\% CI] & NLL reduction [95\% CI] \\
\midrule
All registered wide tables & 18 & 72 & $+1.86\ [ +0.47,+3.63]$ & $0.0431\ [0.0110,0.0841]$ \\
Core width $K>128$ & 9 & 36 & $+3.72\ [ +1.29,+6.63]$ & $0.0862\ [0.0302,0.1541]$ \\
\bottomrule
\end{tabular}
\vspace{2pt}

\parbox{0.98\linewidth}{\footnotesize The panel, folds 1--4, and width grid
$\{16,32,64,96,128,192,256,384,512,768,1024,1536,2048,K\}$ were fixed
before aggregation; the grid is clipped to each support-selected $K$.
All 18 datasets and all 72 folds are retained, including 36 same-width
identity comparisons. Peak memory is measured on the same eight-card host for
both arms; widths are chosen without query labels. Dataset bootstrap uses
100,000 draws and seed 20260920. Bounded-Core candidate-forward time is
$2.76\times$ that of the selected narrower prefix on average across datasets.}
\end{minipage}\end{center}

\begin{center}\begin{minipage}{\linewidth}
\centering
\footnotesize
\setlength{\tabcolsep}{4pt}
\captionof{table}{\textbf{Memory savings at identical selected width, a different comparison.}}
\label{tab:memory-pilot-same-k}
\begin{tabular}{@{}lrrrr@{}}
\toprule
Dataset & $K$ & $\Delta$Acc. pp & $\Delta$NLL & Folded/single peak \\
\midrule
Internet-Ads & 512 & $+0.00$ & $+0.0110$ & 16.3\% \\
Bioresponse & 832 & $+0.00$ & $+0.0081$ & 10.5\% \\
cnae-9 & 256 & $-0.93$ & $-0.0275$ & 35.2\% \\
gina & 448 & $+0.58$ & $-0.0137$ & 18.9\% \\
\bottomrule
\end{tabular}
\vspace{2pt}
\parbox{0.98\linewidth}{\footnotesize Both arms encode exactly the same $K$ columns with query chunk 128. Deltas are folded minus single leaf. Memory is total peak allocation, not the incremental allocation reported in Table~\ref{tab:folding-resource}. Quality is not uniformly improved at identical $K$.}
\end{minipage}\end{center}

Together, these comparisons establish two complementary uses of folding:
encode a fixed Core with less memory, or use a fixed memory budget to encode
a larger Core. The second use closes the loop from systems efficiency to
predictive quality without changing the checkpoint or prediction count.
On TabICLv2, removing the two largest-gain datasets still leaves $+0.75$
accuracy points and a $0.0191$ NLL reduction across the other 16 datasets.
The supplementary numeric records permit reselecting every memory-feasible
prefix and recomputing the paired effects.

\section{Robustness and Resource Analyses}
\label{app:experimental-detail}

Real wide tables vary in how many columns are informative and how many
examples are available. The following interventions change those conditions
one at a time: selected Core width, added irrelevant columns, leaf capacity,
and support context. The final application asks whether the same resource
argument carries over to algorithmically generated features.

\noindent\begin{minipage}{\linewidth}
\centering
\includegraphics[width=0.89\linewidth]{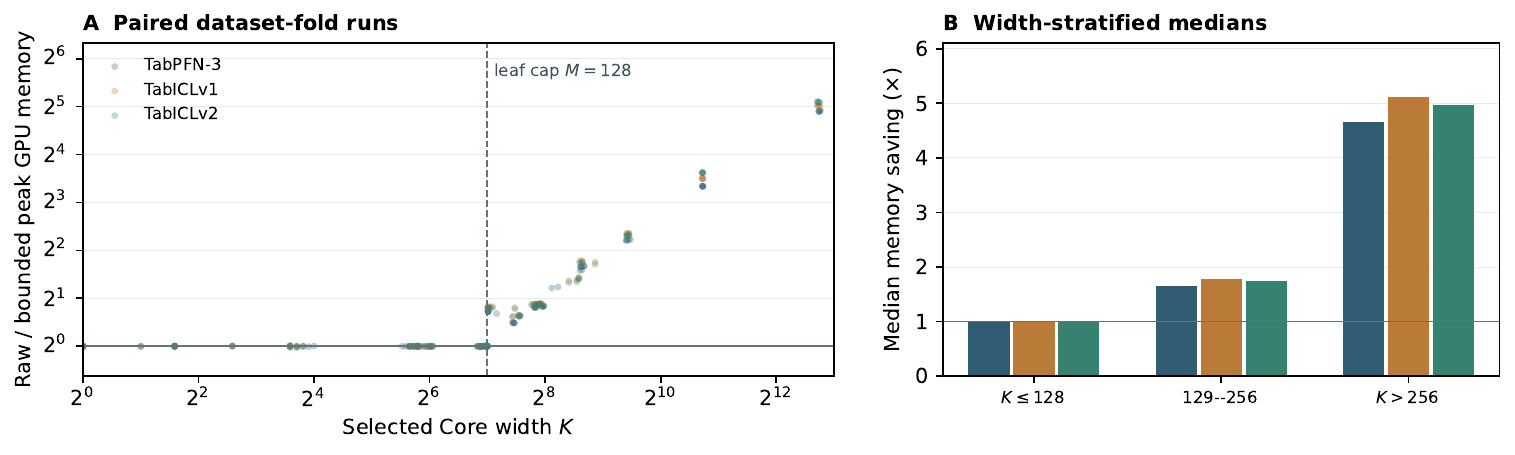}
\captionof{figure}{\textbf{Core memory savings grow with width.} On 90 paired
folds per backbone, median raw/bounded peak-memory ratios for $K>256$ are
$4.66\times$, $5.11\times$, and $4.96\times$ for TabPFN-3, TabICLv1, and
TabICLv2; $K\leq128$ is near parity.}
\label{fig:core-width-resource}
\end{minipage}\par

\subsection{Does folding resist irrelevant width?}

More columns do not necessarily mean more evidence. To separate width from
information, the nuisance intervention adds split-local, label-independent transformations
without deleting them. Doses are nested within seed and the four tasks are
fixed before evaluation. Dose zero has one record per task; each of four
nonzero doses has three seeds per task, giving 52 records. We average seeds
within task and dose and then macro-average tasks. From dose 0 to 512, native
accuracy changes by $-0.605$ points and NLL by $+0.01919$; SCFF changes by
$-0.085$ points and $-0.00105$ NLL. This intervention supports robustness to
the tested nuisance generator and dose range; it does not establish robustness
to arbitrary dependent or adversarial columns.

\noindent\begin{minipage}{\linewidth}
\centering
\includegraphics[width=0.89\linewidth]{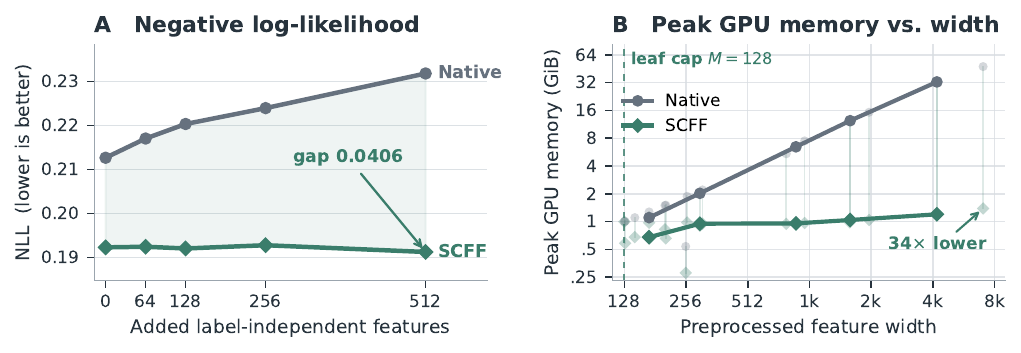}
\captionof{figure}{\textbf{SCFF resists nuisance width while retaining every column.} \textbf{A:} NLL with
0--512 label-independent columns. \textbf{B:} Peak GPU memory as preprocessed
width grows. At 512 added columns, SCFF is $1.440$ accuracy points and
$0.04063$ NLL better than native.}
\label{fig:full-coverage-nuisance}
\end{minipage}\par

\subsection{How do leaf width and context interact?}

The method needs groups small enough to be affordable but large enough to
preserve useful interactions. The leaf ablation changes only $M\in\{32,64,128\}$ over four tasks and five
folds. The router, Tail update, checkpoint, and splits remain fixed. Main Table~\ref{tab:full-coverage-leaf-width} shows that $M=32$ loses
quality, while 64 and 128 both improve over native. The pooled point estimate
at 64 is slightly better in quality, time, and peak memory; $M=128$ remains the
fold-0-selected default used by the full evaluation. We therefore
interpret the sweep as a moderate-width operating range, not evidence that 128
is uniquely optimal.

The context panel asks a different question: whether the effect persists as
the number of support rows changes. One dataset is fixed from each raw-width
band using input metadata, support prefixes are nested, support order and query
rows are fixed, and no quality result enters the context choice. Mixed cells
rule out a universal monotone context law.

\begin{center}
\begin{minipage}{\linewidth}
\centering
\captionof{table}{\textbf{Context-by-width cells.} Width bands and representatives
are fixed from input metadata. Context prefixes reuse the same support order
and query rows within each dataset.}
\label{tab:full-coverage-context-width-detail}
\scriptsize
\setlength{\tabcolsep}{3.2pt}
\resizebox{\linewidth}{!}{%
\begin{tabular}{@{}llrrrrrr@{}}
\toprule
Raw-width band & Dataset & Raw $D$ & Context & $\Delta$Acc. pp & $\Delta$NLL & Peak/native & Time/native \\
\midrule
128-255 & kddcup09\_appetency & 212 & 128 & $+0.000$ & $-0.00791$ & 73.1\% & 0.70$\times$ \\
128-255 & kddcup09\_appetency & 212 & 512 & $+0.000$ & $+0.00408$ & 61.9\% & 0.70$\times$ \\
128-255 & kddcup09\_appetency & 212 & 2048 & $+0.000$ & $-0.00166$ & 54.6\% & 0.78$\times$ \\
256-511 & philippine & 308 & 128 & $-4.102$ & $+0.08142$ & 57.6\% & 0.65$\times$ \\
256-511 & philippine & 308 & 512 & $-2.246$ & $+0.00761$ & 45.4\% & 0.64$\times$ \\
256-511 & philippine & 308 & 2048 & $+0.879$ & $-0.01000$ & 43.0\% & 0.79$\times$ \\
512-2047 & dilbert & 2000 & 128 & $+5.762$ & $-0.18864$ & 12.9\% & 1.63$\times$ \\
512-2047 & dilbert & 2000 & 512 & $+2.148$ & $-0.01944$ & 9.7\% & 1.32$\times$ \\
512-2047 & dilbert & 2000 & 2048 & $-0.098$ & $+0.00049$ & 6.8\% & 0.79$\times$ \\
2048+ & guillermo & 4296 & 128 & $+2.344$ & $-0.08459$ & 6.8\% & 2.41$\times$ \\
2048+ & guillermo & 4296 & 512 & $+8.594$ & $-0.10163$ & 4.6\% & 1.34$\times$ \\
2048+ & guillermo & 4296 & 2048 & $+8.887$ & $-0.11406$ & 3.6\% & 1.05$\times$ \\
\bottomrule
\end{tabular}}
\vspace{2pt}

\parbox{0.98\linewidth}{\scriptsize All 12 cells pass exact feature
coverage, positive leaf-weight, tail-mass, nested-support, fixed-query, and
one-prediction audits. The mixed response of \texttt{dilbert} and the large
gains on \texttt{guillermo} expose context--width heterogeneity rather than
a monotone context-scaling law.}
\end{minipage}
\end{center}

\FloatBarrier

\subsection{Can saved feature memory provide more context?}

Feature width and support length compete for the same device memory. This
experiment spends the saving from narrower feature interactions on more
labeled examples. The separately registered full-coverage mass-Huber folding
study uses an all-feature mass-weighted Huber average, predating the final
residual/prototype update. It is a framework-level resource test, not a
long-context validation of the final Tail update. For each
eligible fold-0 dataset, Native@2,048 defines the measured memory budget.
SCFF support sets are nested prefixes, and $C^\star$ is the largest declared
prefix within 102\% of that budget; quality and latency are hidden until after
memory-only selection. The complete $D\ge2000$, $n\ge10000$ panel includes
four datasets. Three ultra-wide tasks improve substantially, whereas the
2,000-feature boundary \emph{dilbert} slightly worsens. This contrast reveals
the two forces in context reinvestment: more examples can improve inference,
but they must compensate for any representation loss introduced by folding.

\begin{center}\begin{minipage}{\linewidth}
\centering
\captionof{table}{\textbf{Complete fixed-memory study with additional context, including the 2,000-feature boundary.}}
\label{tab:fixed-memory-context-reinvestment-full}
\footnotesize
\setlength{\tabcolsep}{3.0pt}
\begin{tabular}{@{}lrrrrrrr@{}}
\toprule
Dataset & $D$ & Context N / S & Gain & S@2K mem. & S@$C^\star$ mem. & $\Delta$Acc. \% & $\Delta$NLL \\
\midrule
dilbert & 2,000 & 2,048 / 7,680 & 3.75$\times$ & 6.3\% & 20.4\% & $-0.195$ & $+0.00406$ \\
guillermo & 4,296 & 2,048 / 15,360 & 7.50$\times$ & 3.4\% & 24.4\% & $+14.453$ & $-0.22615$ \\
riccardo & 4,296 & 2,048 / 15,360 & 7.50$\times$ & 3.4\% & 24.0\% & $+4.785$ & $-0.14345$ \\
robert & 7,200 & 2,048 / 7,680 & 3.75$\times$ & 3.3\% & 11.4\% & $+8.887$ & $-0.26498$ \\
\midrule
Macro mean & -- & -- & 5.62$\times$ & 4.1\% & 20.1\% & $+6.982$ & $-0.15763$ \\
\bottomrule
\end{tabular}
\vspace{2pt}
\parbox{0.98\linewidth}{\footnotesize All four datasets satisfy the fixed $D\ge2000$, $n\ge10000$ eligibility rule. The main text reports the $D\ge4096$ ultra-wide group; this table includes the complete boundary analysis. All 14 runs completed without fallback or replacement.}
\end{minipage}\end{center}

\subsection{Composing with generated feature spaces}

Feature engineering creates another route to wide tables: a modest raw input
can expand into many derived columns. Holding the generated table fixed lets
us ask whether SCFF helps an ICL model use this larger evidence space more
efficiently, separately from the value of feature generation itself.

The TabPrep application panel contains all 11 classification datasets reported
as improved by tuned TabPrep in its published RealTabPFN-2.5 comparison. This
external opportunity rule is fixed independently of our TabICLv2 outcomes.
Each fold-0 variant shares support/query rows: A uses the raw table, B applies native
inference to every support-fitted generated feature, and C applies SCFF to that
same generated table. C improves NLL on 10 of 11 datasets relative to B while
using 24.3\% of its peak memory and $0.71\times$ its time. However, the macro
C result does not recover the raw-table A baseline. The experiment therefore
supports efficient use of a wide generated feature space, not a
claim that feature generation is always beneficial. Lower NLL is reported as
predictive log-loss improvement, not as a calibration guarantee.

\begin{center}\begin{minipage}{\linewidth}
\centering
\captionof{table}{\textbf{SCFF retains all TabPrep-generated features.} The 11 datasets are fixed by an external TabPrep-positive criterion. A is the raw table, B is native inference on every generated feature, and C applies the selected support-validated Tail update to the same complete generated table. Deltas are dataset-macro right-minus-left changes; W/T/L follows the metric direction.}
\label{tab:full-coverage-tabprep}
\footnotesize
\setlength{\tabcolsep}{3.5pt}
\begin{tabular}{@{}lrrrrr@{}}
\toprule
Comparison & $\Delta$Acc. pp$\uparrow$ & $\Delta$NLL$\downarrow$ & $\Delta\mathrm{AUC}_{\rm OVR}\uparrow$ & Acc. W/T/L & NLL W/T/L \\
\midrule
B $-$ A: generated features & -0.302 & +0.01245 & -0.01065 & 5/1/5 & 3/0/8 \\
C $-$ A: generated + SCFF & -0.239 & +0.00718 & -0.00548 & 4/2/5 & 5/0/6 \\
C $-$ B: generated features + SCFF & +0.064 & -0.00527 & +0.00516 & 4/2/5 & 10/0/1 \\
\bottomrule
\end{tabular}
\vspace{2pt}
\parbox{0.96\linewidth}{\footnotesize Raw widths span 9--170 columns and TabPrep expands them to 136--2,000. Relative to native inference on the generated table, SCFF uses 24.3\% of peak GPU memory and 0.71$\times$ total prediction time. Every generated feature is encoded exactly once; the largest accepted Tail weight is 1.000000.}
\medskip
\resizebox{\linewidth}{!}{%
\begin{tabular}{@{}lrrrrr@{}}
\toprule
Dataset & Raw$\rightarrow$generated & Leaves & $\Delta$Acc. pp & $\Delta$NLL & Peak C/B \\
\midrule
Amazon\_employee\_access & 9$\rightarrow$2,000 & 16 & +0.43 & -0.0159 & 13.7\% \\
APSFailure & 170$\rightarrow$2,000 & 17 & +0.56 & -0.0149 & 41.5\% \\
bank-marketing & 13$\rightarrow$677 & 6 & -0.01 & -0.0013 & 20.5\% \\
churn & 19$\rightarrow$2,000 & 17 & +0.00 & -0.0044 & 7.1\% \\
coil2000\_insurance\_policies & 85$\rightarrow$2,000 & 17 & -0.03 & -0.0048 & 7.2\% \\
Diabetes130US & 47$\rightarrow$2,000 & 17 & -0.02 & -0.0017 & 22.8\% \\
E-CommereShippingData & 10$\rightarrow$959 & 8 & -0.19 & -0.0004 & 13.8\% \\
polish\_companies\_bankruptcy & 64$\rightarrow$2,000 & 17 & +0.05 & -0.0062 & 7.2\% \\
SDSS17 & 11$\rightarrow$2,000 & 17 & +0.15 & -0.0085 & 42.6\% \\
seismic-bumps & 15$\rightarrow$2,000 & 17 & +0.00 & -0.0006 & 7.6\% \\
NATICUSdroid & 86$\rightarrow$136 & 2 & -0.24 & +0.0006 & 92.7\% \\
\bottomrule
\end{tabular}}
\end{minipage}\end{center}

\section{Where the Design Transfers}
\label{app:boundaries}

The experiments point to three conditions for successful transfer: a
compatible representation interface, enough local capacity to encode useful
interactions, and Tail evidence that survives the support check.
SCFF combines aligned leaf outputs before a shared context encoder.
TabPFNv1 is a capacity extension
because its standard inference fixes a 100-feature view. LimiX instead uses the
token-preserving adapter above. The Tail update improves three of five models,
so it does not transfer to every compatible representation interface.

Tail validation currently supports binary tasks; multiclass tasks use Bounded
Core. Stored leaf representations require memory that grows linearly with width, the headline
locked-fold intervals quantify dataset-level uncertainty for the complete
method, and the all-five-fold same-Core intervals are favorable for both metrics
on TabICLv2 but cross zero on TabPFN-3 and TabICLv1. In the secondary folds 1--4
contrast, both intervals also favor bounded folding on TabPFN-3, while those
on TabICLv1 still cross zero. The smaller incremental Tail effect
has favorable intervals for both metrics only on TabICLv2, and the fixed-memory
study uses an earlier Tail update.

\section{Locked-Fold Dataset Effects}
\label{app:heldout-dataset-effects}

The aggregates establish the overall tradeoff; the dataset-level results show
how that tradeoff varies across tasks. The following tables report every
dataset behind the four matched-width locked-fold intervals, so both gains
and regressions can be traced to the same fixed population. Each row is the
mean of folds 1--4; no development-fold observation enters these values.

\begin{center}\begin{minipage}{0.72\linewidth}
\centering
\captionof{table}{\textbf{Locked-fold dataset effects for TabPFN-3.} Each row averages folds 1--4; positive Accuracy and negative NLL favor SCFF.}
\footnotesize\setlength{\tabcolsep}{5pt}
\begin{tabular}{@{}lrr@{}}
\toprule
Dataset (OpenML ID) & $\Delta$Accuracy pp & $\Delta$NLL \\
\midrule
arrhythmia (5) & -0.571 & -0.02489 \\
scene (312) & +0.000 & -0.00043 \\
sylva\_agnostic (1036) & +0.000 & +0.00062 \\
gina\_agnostic (1038) & +2.055 & -0.04217 \\
musk (1116) & +1.440 & -0.12677 \\
gas-drift (1476) & +0.000 & +0.00000 \\
gas-drift-different-concentrations (1477) & -0.146 & +0.00168 \\
christine (41142) & +0.098 & -0.00441 \\
jasmine (41143) & +0.125 & +0.00214 \\
madeline (41144) & +0.637 & -0.00044 \\
philippine (41145) & -0.757 & +0.01648 \\
guillermo (41159) & +4.785 & -0.12165 \\
riccardo (41161) & -0.464 & +0.02530 \\
dilbert (41163) & +1.099 & -0.04067 \\
fabert (41164) & +23.486 & -0.53174 \\
robert (41165) & +3.418 & -0.07173 \\
hiva\_agnostic (46933) & +0.000 & -0.00013 \\
kddcup09\_appetency (46939) & +0.000 & +0.00021 \\
\bottomrule
\end{tabular}
\end{minipage}\end{center}
\clearpage
\begin{center}\begin{minipage}{0.72\linewidth}
\centering
\captionof{table}{\textbf{Locked-fold dataset effects for TabICLv1.} Each row averages folds 1--4; positive Accuracy and negative NLL favor SCFF.}
\footnotesize\setlength{\tabcolsep}{5pt}
\begin{tabular}{@{}lrr@{}}
\toprule
Dataset (OpenML ID) & $\Delta$Accuracy pp & $\Delta$NLL \\
\midrule
arrhythmia (5) & +4.164 & -0.08222 \\
scene (312) & -1.507 & +0.02088 \\
sylva\_agnostic (1036) & +0.488 & -0.00941 \\
gina\_agnostic (1038) & +9.264 & -0.19215 \\
musk (1116) & +0.659 & -0.00627 \\
gas-drift (1476) & -0.024 & +0.00120 \\
gas-drift-different-concentrations (1477) & +0.000 & -0.00301 \\
christine (41142) & +3.320 & -0.03435 \\
jasmine (41143) & +2.011 & -0.01301 \\
madeline (41144) & +31.131 & -0.40828 \\
philippine (41145) & +5.591 & -0.08871 \\
guillermo (41159) & +18.945 & -0.20223 \\
riccardo (41161) & -2.637 & +0.01651 \\
dilbert (41163) & +5.737 & -0.13595 \\
fabert (41164) & +20.605 & -0.42780 \\
robert (41165) & -1.465 & -0.02031 \\
hiva\_agnostic (46933) & +0.000 & -0.00039 \\
kddcup09\_appetency (46939) & -0.024 & -0.00361 \\
\bottomrule
\end{tabular}
\end{minipage}\end{center}
\clearpage
\begin{center}\begin{minipage}{0.72\linewidth}
\centering
\captionof{table}{\textbf{Locked-fold dataset effects for TabICLv2.} Each row averages folds 1--4; positive Accuracy and negative NLL favor SCFF.}
\footnotesize\setlength{\tabcolsep}{5pt}
\begin{tabular}{@{}lrr@{}}
\toprule
Dataset (OpenML ID) & $\Delta$Accuracy pp & $\Delta$NLL \\
\midrule
arrhythmia (5) & +2.485 & -0.02400 \\
scene (312) & +1.143 & -0.02215 \\
sylva\_agnostic (1036) & +0.000 & -0.00023 \\
gina\_agnostic (1038) & +0.865 & -0.01761 \\
musk (1116) & +2.222 & -0.09004 \\
gas-drift (1476) & +0.000 & -0.00012 \\
gas-drift-different-concentrations (1477) & -0.098 & +0.00224 \\
christine (41142) & +0.610 & -0.01382 \\
jasmine (41143) & +0.125 & -0.00066 \\
madeline (41144) & +0.119 & +0.00197 \\
philippine (41145) & +1.123 & -0.01562 \\
guillermo (41159) & +9.814 & -0.12496 \\
riccardo (41161) & +0.684 & -0.03448 \\
dilbert (41163) & +0.391 & -0.01129 \\
fabert (41164) & +1.489 & -0.07060 \\
robert (41165) & +8.032 & -0.19126 \\
hiva\_agnostic (46933) & +0.065 & -0.03949 \\
kddcup09\_appetency (46939) & +0.000 & +0.00351 \\
\bottomrule
\end{tabular}
\end{minipage}\end{center}
\clearpage
\begin{center}\begin{minipage}{0.72\linewidth}
\centering
\captionof{table}{\textbf{Locked-fold dataset effects for RefineICL-24L.} Each row averages folds 1--4; positive Accuracy and negative NLL favor SCFF.}
\footnotesize\setlength{\tabcolsep}{5pt}
\begin{tabular}{@{}lrr@{}}
\toprule
Dataset (OpenML ID) & $\Delta$Accuracy pp & $\Delta$NLL \\
\midrule
arrhythmia (5) & +6.364 & -0.21560 \\
scene (312) & -0.000 & +0.00352 \\
sylva\_agnostic (1036) & -0.049 & +0.00094 \\
gina\_agnostic (1038) & +0.036 & -0.00611 \\
musk (1116) & +3.491 & -0.09720 \\
gas-drift (1476) & -0.049 & +0.00226 \\
gas-drift-different-concentrations (1477) & -0.024 & -0.00120 \\
christine (41142) & +4.077 & -0.13376 \\
jasmine (41143) & -0.545 & +0.01392 \\
madeline (41144) & -4.498 & +0.08619 \\
philippine (41145) & -1.099 & +0.01959 \\
guillermo (41159) & +10.938 & -0.15733 \\
riccardo (41161) & +42.114 & -0.69906 \\
dilbert (41163) & +1.758 & -0.04319 \\
fabert (41164) & -1.733 & +0.01033 \\
robert (41165) & +8.447 & -0.29305 \\
hiva\_agnostic (46933) & +0.000 & -0.00311 \\
kddcup09\_appetency (46939) & +0.000 & +0.00198 \\
\bottomrule
\end{tabular}
\end{minipage}\end{center}
\clearpage

\begin{center}\begin{minipage}{0.92\linewidth}
\centering
\captionof{table}{\textbf{Typical locked-fold dataset effects.} Each entry
first averages folds 1--4 within a dataset. Medians and win/tie/loss counts
describe datasets, not individual query rows; positive Accuracy and negative
NLL favor SCFF.}
\label{tab:heldout-typical-effects}
\footnotesize\setlength{\tabcolsep}{4pt}
\begin{tabular}{@{}lrrrr@{}}
\toprule
Backbone & Median $\Delta$Acc. pp & Acc. W/T/L &
Median $\Delta$NLL & NLL W/T/L \\
\midrule
TabPFN-3 & +0.049 & 9/5/4 & -0.00044 & 11/1/6 \\
TabICLv1 & +1.335 & 11/2/5 & -0.01666 & 15/0/3 \\
TabICLv2 & +0.647 & 14/3/1 & -0.01662 & 15/0/3 \\
RefineICL-24L & 0.000 & 8/3/7 & -0.00216 & 10/0/8 \\
\bottomrule
\end{tabular}
\end{minipage}\end{center}

\section{Reproducibility}

The reproduction materials follow the paper's comparisons: complete-method
quality, fixed-Core folding, incremental Tail recovery, and resource
reinvestment. Each table can be traced to its paired records and regenerated
under the stated inference contract.

The final residual/prototype evaluation is identified by
\nolinkurl{scff_core_raw_tail_signal_optimization_v1}; the separately labeled
fixed-memory experiment uses the all-feature mass-weighted Huber average described
above. Reproduction requires the checkpoint hashes, OpenML IDs, fold
records, support/query hashes, preprocessing records, method and adapter
revisions, and runtime fingerprints. Paired fold-level metrics, failure
histories, and table-generation scripts are released without dataset rows or
per-query predictions. The appendix reports results from
the final method and the separately labeled fixed-memory study.

Within the supplementary archive, the portable locked-fold metric root is
\nolinkurl{paper_evidence/artifacts/scff_paper/core_raw_tail_signal_v1/};
\nolinkurl{full_method_all_five_folds.json} indexes the paired
fold-level metrics and \nolinkurl{tail_confirmation_folds1_4.json} summarizes
the locked Tail comparison. \nolinkurl{core_confirmation_bootstrap_receipt.json} records the
same-Core dataset-bootstrap seed, sample count, estimands, and intervals. The scripts
\nolinkurl{render_scff_heldout_full_method_table.py} and
\nolinkurl{render_scff_influence_sensitivity.py} verify coverage and source
hashes before regenerating the reported tables.
The RefineICL-24L row has a paired metric record, checkpoint digest, and
separate anonymous training-source archive here. Its exact concurrent
checkpoint is not redistributed; independent re-execution of that backbone
awaits the companion weights.

\fi

\end{document}